\documentclass[10pt, a4paper, logo]{googledeepmind} %
\usepackage{times}

\usepackage{hyperref}
\usepackage[leftcaption]{sidecap} % caption 在右边

\usepackage{amsmath,amsfonts,bm}
\usepackage{natbib}
\usepackage{fancyhdr}
\def\eqref#1{equation~\ref{#1}}
\def\1{\bm{1}}

\def\vb{{\bm{b}}}

\def\vh{{\bm{h}}}

\def\vx{{\bm{x}}}
\def\vy{{\bm{y}}}

\def\vH{{\bm{H}}}

\def\mH{{\bm{H}}}

\DeclareMathAlphabet{\mathsfit}{\encodingdefault}{\sfdefault}{m}{sl}
\SetMathAlphabet{\mathsfit}{bold}{\encodingdefault}{\sfdefault}{bx}{n}

\def\gD{{\mathcal{D}}}

\def\gL{{\mathcal{L}}}

\def\gX{{\mathcal{X}}}
\def\gY{{\mathcal{Y}}}

\def\bbE{{\mathbb{E}}}

\def\sR{{\mathbb{R}}}

\usepackage{hyperref}
\usepackage{url}
\usepackage{booktabs}
\usepackage{graphicx}
\usepackage{subcaption}
\usepackage{amssymb}
\usepackage[most]{tcolorbox}
\usepackage[table]{xcolor}
\tcbuselibrary{breakable} % 加载分页支持库
\usepackage{amsmath} 
\usepackage{siunitx}
\usepackage{colortbl}

\usepackage{microtype}
\usepackage{color}
\usepackage{wrapfig}
\usepackage{natbib}
\usepackage{colortbl}
\usepackage{microtype}
\usepackage{graphicx}
\usepackage{booktabs} % for professional tables
\usepackage{array}
\usepackage{textcomp}
\usepackage{stfloats}
\usepackage{float}
\usepackage{verbatim}
\usepackage[ruled, linesnumbered, lined]{algorithm2e}
\usepackage{multirow}
\usepackage{enumitem}

\definecolor{BestColor}{HTML}{C8E6C9}  % 一个柔和的绿色
\definecolor{SecondBestColor}{HTML}{FFF9C4} % 一个非常淡的黄色

\usepackage{tcolorbox}
\usepackage{amsmath,amsfonts}
\def\rre{\textcolor{red}}
\definecolor{ggg}{RGB}{26,179,0}
\definecolor{rrr}{RGB}{179,0,0}
\definecolor{oodc}{RGB}{31,73,121}
\definecolor{idc}{RGB}{68,142,68}

\definecolor{mygray}{gray}{0.9}

\newcommand\blfootnote[1]{%
  \begingroup
  \renewcommand\@makefnmark{}%
  \footnotetext{#1}%
  \endgroup
}

\def\Pref#1#2{\bm{1}_{{#1} \succ {#2}}}
\def\Bias#1#2{\bm{b}}

\def\MI#1#2{{I}\left(#1;#2\right)}

\newtcolorbox{examplebox}[2][]{ % 允许传入可选参数 [#1] 和必选标题参数 {#2}
    breakable, % 关键：允许跨页分割
    enhanced, % 增强模式（可选，支持更多样式）
    colback=white, % 框体内背景色
    colframe=cyan, % 边框颜色
    coltitle=white, % 标题文字颜色
    fonttitle=\bfseries, % 标题字体加粗
    title=#2, % 框体标题（第二个必选参数）
    overlay middle={\draw[cyan, line width=1pt](frame.south west)--(frame.south east);}, % 分割处添加横线
    overlay last={\draw[cyan, line width=1pt](frame.south west)--(frame.south east);}, % 最后一页底部横线
    #1 % 允许在调用时传入其他可选参数以覆盖默认样式
}

\usepackage[T1]{fontenc}
\usepackage{booktabs}      % 用于漂亮的表格线 (toprule, midrule, etc.)
\usepackage{graphicx}      % 用于 \resizebox
\usepackage[table]{xcolor} % 用于颜色
\usepackage{siunitx}       % 用于 S 列，实现小数点对齐
\usepackage{etoolbox}      % 用于 \ifstrequal，实现条件判断
\usepackage[normalem]{ulem}     % 用于更灵活的下划线 \uline

\definecolor{impcolor}{HTML}{2E8B57} % 提升使用的海绿色 (SeaGreen)

\newcommand{\improvementstyle}[1]{$^{\textcolor{impcolor}{\tiny #1}}$}

\newcommand{\scoreimp}[2]{%
  \textbf{#1}%
  \ifstrequal{#2}{+0.0}{}{%
    \ifstrequal{#2}{0.0}{}{%
      \makebox[0pt][l]{\improvementstyle{#2}}%
    }%
  }%
}

\title{Eliminating Inductive Bias in Reward Models with Information-Theoretic Guidance}

\author[1,2,3]{Zhuo Li$^{*}$}
\author[1]{Pengyu Cheng\textsuperscript{\dag}}
\author[1]{Zhechao Yu}
\author[1]{Feifei Tong}
\author[3]{Anningzhe Gao}
\author[2,3]{Tsung-Hui Chang}
\author[3]{Xiang Wan}
\author[1]{Erchao Zhao}
\author[1]{Xiaoxi Jiang}
\author[1]{Guanjun Jiang}
\affil[1]{Qwen Large Model Application Team, Alibaba}
\affil[2]{The Chinese University of Hong Kong, Shenzhen}
\affil[3]{Shenzhen Research Institute of Big Data}

\begin{abstract}
Reward models (RMs) are essential in reinforcement learning from human feedback (RLHF) to align large language models (LLMs) with human values. However, RM training data is commonly recognized as low-quality, containing inductive biases that can easily lead to overfitting and reward hacking. For example, more detailed and comprehensive responses are usually human-preferred but with more words, leading response length to become one of the inevitable inductive biases. A limited number of prior RM debiasing approaches either target a single specific type of bias or model the problem with only simple linear correlations, \textit{e.g.},  Pearson coefficients. To mitigate more complex and diverse inductive biases in reward modeling, we introduce a novel information-theoretic debiasing method called \textbf{D}ebiasing via \textbf{I}nformation optimization for \textbf{R}M (DIR). Inspired by the information bottleneck (IB), we maximize the mutual information (MI) between RM scores and human preference pairs, while minimizing the MI between RM outputs and biased attributes of preference inputs. With theoretical justification from information theory, DIR can handle more sophisticated types of biases with non-linear correlations, broadly extending the real-world application scenarios for RM debiasing methods. In experiments, we verify the effectiveness of DIR with three types of inductive biases: \textit{response length}, \textit{sycophancy}, and \textit{format}.  We discover that DIR not only effectively mitigates target inductive biases but also enhances RLHF performance across diverse benchmarks, yielding better generalization abilities. The code and training recipes are available at \url{https://github.com/Qwen-Applications/DIR}.
\end{abstract}
\begin{document}
\maketitle
%\vspace{-1mm}
\section{Introduction}
%\vspace{-1mm}
Aligning Large Language Models (LLMs)~\citep{openai2024gpt4technicalreport, touvron2023llamaopenefficientfoundation,yang2024qwen2} with human values is a fundamental technique to guarantee the helpfulness and harmlessness of LLM responses, which has been widely applied in various open-domain conversational scenarios~\citep{ouyang2022training,team2025kimi, touvron2023llamaopenefficientfoundation,comanici2025gemini25pushingfrontier,openai2024gpt4technicalreport}. Toward more human-preferred LLM behaviors, reinforcement learning from human feedback (RLHF)~\citep{ouyang2022training,rafailov2024direct,deepseekai2025deepseekr1incentivizingreasoningcapability} has become the mainstream approach, which first trains a reward model (RM) on a collection of human preference response pairs, then scores the LLM's responses with the learned RM as the rewards to conduct a reinforcement learning (RL)~\citep{ouyang2022training}.
% Reinforcement Learning from Human Feedback (RLHF)~\citep{ouyang2022training,rafailov2024direct,deepseekai2025deepseekr1incentivizingreasoningcapability} has become the fundamental technique for aligning Large Language Models (LLMs)~\citep{openai2024gpt4technicalreport, touvron2023llamaopenefficientfoundation,yang2024qwen2} with human preferences. 
% \pengyu{add one sentence describing how RLHF works, how RM learns, RLHF learns}. 
%However, the stability and effectiveness of RLHF have been continuously challenged by reward hacking, a phenomenon where the policy model exploits loopholes in a proxy reward model (RM) to achieve high scores without fulfilling the intended human objectives
%
%
Although dominantly applied in LLM post-training~\citep{deepseekai2025deepseekr1incentivizingreasoningcapability,touvron2023llamaopenefficientfoundation}, RLHF has continuously been criticized by its training instability, which can easily lead to LLM's training collapse and overfitting~\citep{rafailov2024direct,zhu2024iterative,yu2025dapo}. 

Among the many factors leading to the training instability of RLHF, the issue of reward model hacking is non-negligible: due to the low quality of human preferences \citep{zeng2024diversified,liu2025skywork,wang2025helpsteer3preferenceopenhumanannotatedpreference,liu2024skyworkrewardbagtricksreward}, which contains massive preference conflicts and inductive biases, the reward model can easily be mislead by irrelative data attributes instead of targeting on the real content quality~\citep{skalse2025definingcharacterizingrewardhacking, gao2023scaling,amodei2016concreteproblemsaisafety}. For example, annotators are always instructed to choose more informative responses, whereas more detailed responses usually have longer response lengths. Learning on this biased human feedback dataset can lead the reward model to ignore the true response quality and only favor responses with longer lengths~\citep{singhal2023long}. Besides the response length bias, stylistic and format patterns~\citep{zhang-etal-2025-lists}, and sycophantic phrasing~\citep{sharma2023towards,denison2024sycophancy} have also been recently recognized as the typical \textit{inductive biases} in reward modeling, which are unrelated to response quality but strongly correlated with the preference annotations~\citep{sharma2023towards,liu2024rmbenchbenchmarkingrewardmodels}. Learning with inductive biases can easily disrupt RMs' learning targets and critically hinders the reliability and the generalization ability of RLHF~\citep{gao2023scaling,coste2023reward}. 

%These attributes, which are unrelated to response quality but strongly correlated with the preference annotations,  and are commonly referred to as \textit{inductive bias}\pengyu{reference}. 

%The vulnerabilities of RMs are commonly derived from the low quality of the human feedback annotation, which contains various toxic \textit{inductive biases}. 

To mitigate inductive biases in reward modeling, a limited number of recent studies have made some preliminary explorations. \citet{bu2025beyond,chen2024odindisentangledrewardmitigates,zhang-etal-2025-lists} consider Pearson Coefficient~\citep{benesty2009pearson} as the bias measurement, which is then jointly minimized with the reward modeling loss. However, the Pearson Coefficient only captures the simplest \textit{linear} correlation between RM and the bias attributes, which are not sufficiently applicable in more general scenarios. \citet{shen2023loose} adds another RM head to predict response length score, which is only applicable with scalar types of inductive biases and lacks theoretical justification. 
 %
%While various methods~\citep{singhal2023long,sharma2023towards,zhang-etal-2025-lists,shen2023loose,chen2024odindisentangledrewardmitigates,bu2025beyond,wang2025beyond} have been proposed to mitigate these inductive bias, they fall into a dilemma of being either too specific or fundamentally misguided in their debiasing strategy. Heuristic approaches like PoE~\citep{shen2023loose}, ALBM~\citep{bu2025beyond} and LE~\citep{zhang-etal-2025-lists}, for instance, are often limited to a single bias type (e.g., length and format) and rely on oversimplified assumptions, thus lacking generality. More principled frameworks face a deeper, methodological trade-off.
%
%\citet{wang2025beyond} imposes overly-restrictive external constraints, such as enforcing distributional invariance via the  Maximum Mean Discrepancy (MMD)~\citep{JMLR:v13:gretton12a}, risk distorting the reward landscape by collapsing the scores of functionally disparate response groups. 
%
\citet{wang2025beyond} introduces overly restrictive external constraints, such as enforcing distributional invariance by minizmizing the Maximum Mean Discrepancy (MMD)~\citep{JMLR:v13:gretton12a} between the chosen and rejected distributions. This approach risks distorting the reward landscape by inducing a collapse in the scores of functionally distinct response groups.
%
%On the other hand, approaches relying on unreliable indirect internal compression, like the standard information bottleneck~\citep{tishby2000informationbottleneckmethod} used in InfoRM~\citep{miao2024inform}, offer no guarantee of discarding a bias attribute, especially when it is strongly correlated with the true preference signal.
%
In contrast, methods utilizing general compression strategies, such as the Information Bottleneck~\citep{tishby2000informationbottleneckmethod} used in InfoRM~\citep{miao2024inform}, cannot guarantee the mitigation of inductive biases, since no explicit constraint is applied to the biased attributes within the optimization.  % are strongly correlated with the true preference signal.

% To address the inductive biases of RM generally with theoretical guarantees, we proposed an information-theoretic bias-disentangling framework called \textbf{D}ebiasing via \textbf{I}nformation optimization for \textbf{R}M (DIR). More specifically, we model the complicated inductive biases in human feedback with the concept of mutual information (MI) from the perspective of information theory~\citep{kullback1997information}. Then we maximize the mutual information between the RM prediction and the preference data inputs, and minimize the mutual information between the bias attributes and the RM scores, simultaneously.
To uniformly eliminate inductive biases from reward modeling with theoretical guarantees, we propose an \textit{information-theoretic} debiasing framework called \textbf{D}ebiasing via \textbf{I}nformation optimization for \textbf{R}Ms (DIR). %as illustrated in Figure \ref{fig:framework}. 
Inspired by the information bottleneck methods~\citep{tishby2000informationbottleneckmethod}, we model the complicated inductive biases between irrelevant attributes and human preferences by the concept of mutual information (MI) from the perspective of information theory~\citep{kullback1997information}. %Our framework then simultaneously optimizes two competing objectives: DIR maximizes the mutual information between the comparative representation and the true preference label, while concurrently minimizing the mutual information between the representation and the relevant bias attributes.
The proposed debiasing method maximizes the mutual information between the content quality of response pairs and the ground-truth preference label, while simultaneously minimizing the mutual information between the RM's preference prediction and the irrelevant bias attributes.
To tackle the intractable MI calculation~\citep{poole2019variational}, we estimated the above objective with two variational bounds: the Barber-Agakov (BA) lower bound~\citep{barber2004algorithm} for the MI maximization and the contrastive log-ratio upper bound (CLUB)~\citep{cheng2020club} for the MI minimization. 
%a variational lower bound preserves essential preference information, while another variational upper bound actively suppresses information related to the bias. 
%
To further extend the method's applicability, we design a comparative regularizer that operates on relative bias attributes between response pairs, rather than on the isolated attributes of individual responses.
%we design a comparative regularizer that operates on relative bias attributes between response pairs, rather than absolute values,
This modification allows DIR to generally handle diverse and complex types of biases without distorting the underlying reward landscape, yielding broader application scenarios. %a debiased reward model that demonstrates superior generalization and performance in downstream RLHF tasks. 
% 
% Extensive experiments are conducted on both LLM's capability benchmark (e.g., GSM8K, Hellaswag, MMLU, ArenaHard and MT-Bench) and Reward Model benchmarks (e.g., RM-Bench and RewardBench) across different bias settings (length, format and sycophancy), demonstrating that our DIR method significantly outperforms existing debiasing baselines and leads to more robust and reliable LLM alignments.
We conduct extensive experiments on both LLM capability benchmarks (\textit{e.g.}, GSM8K~\citep{cobbe2021trainingverifierssolvemath}, MMLU~\citep{hendrycks2021measuringmassivemultitasklanguage}, ArenaHard~\citep{li2024crowdsourceddatahighqualitybenchmarks}, and MT-Bench~\citep{zheng2023judgingllmasajudgemtbenchchatbot}) and reward model benchmarks (\textit{e.g.}, RM-Bench~\citep{liu2024rmbenchbenchmarkingrewardmodels} and RewardBench~\citep{lambert2024rewardbenchevaluatingrewardmodels}), under multiple bias settings including response length, formatting style, and sycophancy. The numerical results demonstrate that the proposed DIR method consistently outperforms existing debiasing baselines, yielding more robust and reliable alignment of LLMs.

%which allows DIR to robustly handle diverse and complex biases without distorting the underlying reward landscape, leading to a debiased RM that demonstrates better generalization and performance in downstream RLHF tasks.
% We summarize our contributions as follows:
% \begin{itemize}[leftmargin=*,align=left]
% \vspace{-2mm}
% \item We propose a novel, explicit RM debiasing framework information-theoretic framework that transforms debiasing from an indirect hope into a direct, supervised optimization objective, offering a more principled and targeted solution.
% %We propose a novel, explicit information-theoretic framework that transforms debiasing from an indirect hope into a direct, supervised optimization objective, offering a more principled and targeted solution.
% \item We design a practical and generalizable implementation that is computationally efficient and can be seamlessly adapted to mitigate diverse forms of bias without requiring architectural modifications to the base RM.
% \item 
% \end{itemize}

%\vspace{-1mm}
\section{Preliminary}
%\vspace{-1mm}
\paragraph{Reinforcement Learning from Human Feedback (RLHF)} has become one of the essential training processes to align LLMs with human values~\citep{ouyang2022traininglanguagemodelsfollow}. With a well-learned reward model (RM) $r_\phi(\vx,\vy)$ scoring the degree of human preference of generated response $\vy \in \gY$ given input prompt $\vx \in \gX$, RLHF optimizes the LLM policy $\pi_\theta(\vy|\vx)$  with the follow objective: 
\begin{equation}\label{eq:RLHF}
    \mathbb{E}_{\vx\sim \mathcal{X},\vy\sim\pi_\theta(\cdot|\vx)} [ r(\vx, \vy) - \beta\cdot\text{KL}[\pi_\theta(\vy|\vx)\Vert\pi_\text{ref}(\vy|\vx)]],
\end{equation}
where $\pi_\text{ref}(\vy | \vx)$ is the initial model policy served as a reference, $\beta > 0$ controls the strength of a Kullback-Leibler (KL) divergence~\citep{csiszar1975divergence} between the reference model $\pi_\text{ref}(\vy|\vx)$ and the learning policy $\pi_\theta(\vy|\vx)$. 
%
%for each input query $\vx$ by maximizing the following objective on training samples $\mathcal{D}=\{\vx_1,\vx_2,...,\vx_M\}$, while not deviating too strongly from the initial distribution:
%
 To train LLMs with the above objective, Proximal Policy Optimization (PPO) \citep{schulman2017proximalpolicyoptimizationalgorithms} has been recognized as the mainstream optimization approach~\citep{openai2024gpt4technicalreport,bai2023qwentechnicalreport,rafailov2024direct}. Group Relative Policy Optimization (GRPO)~\citep{shao2024deepseekmathpushinglimitsmathematical} further removes the critic model in PPO and uses a simplified group-related advantage approximation instead, which has shown competitive performance with practically simpler infrastructures~\citep{deepseekai2025deepseekr1incentivizingreasoningcapability,yang2025qwen3technicalreport}.

\paragraph{Reward Modeling} targets learning the human preference distribution via a parameterized reward model (RM) $r_\phi: \gX \times \gY \rightarrow \sR $, where $r_\phi(\vx,\vy)$ is the predicted reward score of the input prompt $\vx$ and the corresponding response $\vy$. For every input $\vx$, given a pair of response $(\vy, \bar{\vy})$, we can calculate the ``preference'' by comparing the reward scores: if $r(\vx, \vy) > r(\vx, \bar{\vy})$, then $\vy$ is predicted as a more ``preferred'' response than $\bar{\vy}$ (denote as $\vy \succ \bar{\vy}$) and vice versa. We use a binary indicator $\Pref{\vy}{\bar{\vy}}$ to represent the event of ``human preference'': $\Pref{\vy}{\bar{\vy}} = 1$, if $\vy \succ \bar{\vy}$; and $\Pref{\vy}{ \bar{\vy}} = 0 $, if $\vy \prec \bar{\vy}$. 
% \begin{equation}
% \Pref{\vy}{\bar{\vy}} = 1, \text{ if } \vy \succ \bar{\vy}, \text{ and } 
% \end{equation}
Then the RM predicting  preference $\Pref{\vy}{\bar{\vy}}$ can be regarded as drawing a conditional Bernoulli~\citep{chen1997statistical} random variable from:
\begin{align}\label{eq:rm_predicted_probability}
  q_{\phi}\Big(\Pref{\vy}{ \bar{\vy}} = 1 \Big|\vx, \vy, \bar{\vy}\Big) = \frac{\exp(r_\phi(\vx, \vy))}{\exp(r_\phi(\vx, \vy)) + \exp(r_\phi(\vx, \bar{\vy}))}  = \sigma\Big(r_\phi(\vx,\vy) - r_\phi(\vx, \bar{\vy})\Big),
\end{align}
where $\sigma(\cdot)$ is a Sigmoid function. Note that the ground-truth human preference distribution $p^*(\Pref{\vy}{ \bar{\vy}} | \vx, \vy, \bar{\vy})$ is unknown. To optimize the reward model, we instead maximize the log-likelihood of $q_\phi$ with a group of human preference data $\gD_\text{Pref} = \{(\vx_i, \vy_i^w, \vy_i^l)\}_{i=1}^N$:
\begin{align}
\gL_\text{RM}(\phi) = & - \bbE_{\Pref{\vy}{ \bar{\vy}} \sim p^*} \left[ \log q_\phi \Big( \Pref{\vy}{ \bar{\vy}} \Big| \vx,\vy,\bar{\vy} \Big) \right]  \approx - \frac{1}{N} \sum_{i=1}^N \left[ \log q_\phi( \vy_i^w \succ \vy_i^l | \vx_i, \vy_i^w, \vy_i^l)\right] \nonumber \\ = & - \frac{1}{N} \sum_{i=1}^N [\log \sigma(r_\phi(\vx_i,\vy_i^w) - r_\phi(\vx_i, \vy_i^l))], \label{eq:rm-bt-loss} \text{\ \ \ \ \ \ \ \ \  (by \eqref{eq:rm_predicted_probability})}
\end{align}
where each $\vy^w \succ \vy^l$ is annotated by human judgment with respect to the response content quality.
Equation~\ref{eq:rm-bt-loss} is commonly recognized as the Bradley-Terry ranking loss~\citep{bradley1952rank}.

\paragraph{Information-theoretic Methods} optimize deep models from the perspective of information theory~\citep{chen2016infogan,hjelm2019learningdeeprepresentationsmutual,
yuanimproving,cheng2021fairfilcontrastiveneuraldebiasing}. The core methodology of information-theoretic methods is modeling the feed-forward process of neural networks as an information channel transmission, where the correlation between different neural embeddings is measured by mutual information (MI) as:
\begin{equation}\label{eq:mi}
    \MI{\vx}{\vy}=\mathbb{E}_{p(\vx,\vy)}\Big[ \log\frac{p(\vx,\vy)}{p(\vx)p(\vy)}\Big] = \text{KL} \Big[p(\vx,\vy) \Vert p(\vx) p(\vy)\Big],
\end{equation}
where $p(\vx,\vy)$ is the joint distribution, and $p(\vx)$ and $p(\vy)$ are the marginal distributions. 
%
%The mutual information $\MI{\vx}{\vy}$ measures the correlation between the two variables $\vx$ and $\vy$ as the Kullback-Leibler~\citep{kullback1997information} (KL) divergence between the joint distribution $p(\vx,\vy)$ and the product of marginal distributions $p(\vx)p(\vy)$.
%
Due to its general ability to capture arbitrary non-linear correlations, MI has achieved considerable success as a learning objective in various deep learning tasks~\citep{chen2016infogan,belghazi2018mutual,hjelm2019learningdeeprepresentationsmutual}.
However, due to the intractable expectation \textit{w.r.t.} $p(\vx,\vy)$, the exact MI value in \eqref{eq:mi} is challenging to compute, especially when only samples from $p(\vx,\vy)$ are provided. To address this, several approximation methods have been proposed to estimate MI from samples using tractable variational bounds~\citep{oord2018representation,cheng2020club,belghazi2021minemutualinformationneural}. Barber-Agakov (BA) bound~\citep{barber2004algorithm} provides a simple lower bound approximation of MI, by introducing a variational approximation $q_\theta(\vy|\vx)$: 
\begin{align}\label{eq:ba_bound}
    \MI{\vx}{\vy}  \geq  \bbE_{p(\vx,\vy)} [\log q_\theta(\vy|\vx)] + H[p] = : I_\text{BA}(\vx;\vy),
\end{align}
where $H[p]$ is the entropy of the ground-truth distribution $p(\vx,\vy)$. 
%For {MI maximization}, a common strategy is to optimize an achievable variational {lower bound}.
%
%Conversely, for {MI minimization}, one must optimize an {upper bound}, where a widely used example is the Contrastive Log-ratio Upper Bound (CLUB)~\citep{cheng2020club} that approximates the conditional distribution $p(\vy|\vx)$ with a variational network $q_\phi(\vy|\vx)$ and is defined as:
Besides, \citet{cheng2020club} propose a variational contrastive log-ratio upper bound (CLUB) also utilizing the variational approximation $q_\theta(\vy|\vx)$:
\begin{equation}\label{eq:club}
   \MI{\vx}{\vy} \leq  \bbE_{p(\vx,\vy)} [ \log q_\theta(\vy|\vx)]  - \bbE_{p(\vx)p(\vy)} [\log q_\theta(\vy|\vx) ] = : {I}_{\text{CLUB}}(\vx;\vy).
\end{equation}
By minimizing \eqref{eq:club}, the amount of information between $\vx$ and $\vy$ can be effectively reduced. We provide the proof of BA bound and CLUB in Appendix~\ref{app:proof_ba}\&~\ref{app:proof_club}. A well-known application of information-theoretic methods is the \textit{information bottleneck} (IB)~\citep{tishby2000informationbottleneckmethod}, which aims to learn a compressed but informative representation $\vh$ of an input $\vx$ to the output $\vy$ as a trade-off between two MI terms:
\begin{equation}\label{eq:infor_b}
\min_{\vh} \MI{\vx}{\vh} - \lambda \cdot \MI{\vh}{\vy},
\end{equation}
%where one seeks to minimize $\mathcal{L}_{\text{IB}}$ by finding an optimal encoding function $p(\vh|\vx)$.
where hyper-parameter $\lambda > 0$ controls the balance between compressing the input $\vx$  and retaining relevant information for the prediction $\vy$. IB has been recognized as a powerful tool for representation learning and widely applied to diverse deep learning scenarios~\citep{saxe2019information,wan2021multi,federici2020learning}. %Optimizing this objective naturally requires the MI estimators discussed previously, making it a powerful conceptual tool for representation learning.

\section{Methodology}
We begin by revisiting reward modeling from a perspective of information theory. Given an input query $\vx\in \gX$ and a pair of responses $\vy,\bar{\vy} \in \gY$, we denote $\vb$ as a concerned bias attribute with respect to $(\vx,\vy, \bar{\vy})$. Our debiasing target is to learn a reward model $r_{\phi}(\vx, \vy)$ that produces predictions $\Pref{\vy}{\bar{\vy}}$ highly correlated with the content quality of response pairs $(\vy,\bar{\vy})$ while eliminating any indication of the pre-defined bias attribute $\vb$.
Motivated by the information bottleneck method in \eqref{eq:infor_b}, we model the debiasing objective as maximizing the mutual information between the input response content and the RM preference prediction, while minimizing the mutual information between the RM prediction and the bias attribute: %l Formally, we measure the correlation between RM predictions and ground-truths human define our objective as follows:
\begin{equation}\label{eq:surrogate_objective}
 \max_\phi \underbrace{\MI{\Pref{\vy}{ \bar{\vy}}}{\vx, \vy, \bar{\vy}}}_{\text{Preference Term}} - \lambda \cdot \underbrace{\MI{\Pref{\vy}{ \bar{\vy}}}{\Bias{\vy}{\bar{\vy}}}}_{\text{Debiasing Term}},
\end{equation}
where $\lambda>0$ is a hyper-parameter balancing the trade-off between preference learning and debiasing. Ideally, minimizing \eqref{eq:surrogate_objective} should encourage the reward model $r_\phi$ to capture the true performance signal from the input triplet $(\vx,\vy,\bar{\vy})$, while decreasing the reliance on the bias attribute $\vb$. 

However, directly optimizing the mutual information-based objective is computationally intractable due to the difficulty in estimating mutual information in high-dimensional spaces~\citep{poole2019variational}. Hence, we follow the prior works~\citep{oord2018representation,cheng2021fairfilcontrastiveneuraldebiasing} and utilize the variational mutual information bounds (as in equation \ref{eq:ba_bound} \& \ref{eq:club}) to estimate the preference term and debiasing term of \eqref{eq:surrogate_objective}, separately.

%\subsection{Practical Implementation with Differentiable Losses}
%To render~\eqref{eq:surrogate_objective} tractable, we derive differentiable surrogate losses that approximate its constituent terms. Specifically, we minimize a total loss $\mathcal{L}_{\text{total}}$ composed of a preference loss $\mathcal{L}_{\text{Preference}}$ and a debiasing loss $\mathcal{L}_{\text{Debiasing}}$.

\paragraph{Preference Term Estimation. } Instead of directly enlarging $\MI{\Pref{\vy}{ \bar{\vy}}}{\vx, \vy, \bar{\vy}}$, we can maximize its lower bound approximation by applying the BA estimator as in \eqref{eq:ba_bound}:
\begin{align}\label{eq:ba_bound_of_dir}
 \MI{\Pref{\vy}{ \vy}}{\vx, \vy, \bar{\vy}} \geq \bbE_{p^*(\vx, \vy, \bar{\vy}, \Pref{\vy}{\bar{\vy}})} [\log q_\phi(\Pref{\vy}{\bar{\vy}} | \vx,\vy,\bar{\vy})] + H[p^*],
\end{align}
where $p^*(\vx, \vy, \bar{\vy}, \Pref{\vy}{\bar{\vy}})$ is the ground-truth joint distribution of human preference training data, and $H[p^*]$ is the entropy of the data distribution $p^*$ as a constant to the learning parameters. %the ground-truth human preference distribution $p^*$. 
By \eqref{eq:rm-bt-loss}, the expectation term in the right-hand side of \eqref{eq:ba_bound_of_dir} is exactly the commonly used Bradley-Terry ranking loss~\citep{bradley1952rank} of reward modeling~\citep{azar2024general,cheng2024adversarial}. Hence, minimizing the RM ranking loss actually maximizes the mutual information between the preference prediction $\Pref{\vy}{\bar{\vy}}$ and the input triplet $(\vx,\vy,\bar{\vy})$, encouraging the reward model $r_\phi$ to output a higher score to the preferred response $\vy$. Therefore, given a batch of preference data $\gD_\text{Pref} = \{(\vx_i, \vy_i^w, \vy_i^l) | \vy_i^w \succ \vy_i^l \}_{i=1}^B$, we maximize the following RM ranking loss to approximate the preference term in \eqref{eq:surrogate_objective} instead: 
\begin{equation}
    \mathcal{L}_{\text{Preference}}(\phi):= - \frac{1}{B} \sum_{i=1}^B \Big[\log \sigma(r_\phi(\vx_i,\vy_i^w) - r_\phi(\vx_i. \vy_i^l))\Big].
\end{equation}
\paragraph{Debiasing Term Estimation.}
Since the response pairs $(\vx, \vy,\bar{\vy})$ contains sufficient information to determine the bias attribute $\Bias{\vy}{\bar{\vy}}$, we can conclude that the RM forward process ($\Bias{\vy}{\hat{\vy}} \rightarrow (\vx,\vy,\bar{\vy}) \rightarrow \mH \rightarrow \Pref{\vy}{\bar{\vy}}$) is a Markov Chain~\citep{shannon1948mathematical}, where $\mH = [\vh_\phi(\vx, \vy), \vh_\phi(\vx,\bar{\vy})]$ is the last hidden states of the RM's transformer backbone. According to the data processing inequality~\citep{shannon1948mathematical} and the CLUB upper bound~\citep{cheng2020club} in \eqref{eq:club}, we have
\begin{equation}\label{eq:club_debias_obj}
    \MI{\Pref{\vy}{\bar{\vy}}}{\Bias{\vy}{\hat{\vy}}} \leq  \MI{\mH}{\Bias{\vy}{\hat{\vy}}} \leq I_\text{CLUB} (\mH ;\Bias{\vy}{\hat{\vy}}),
\end{equation}
where $I_\text{CLUB} (\mH ;\Bias{\vy}{\hat{\vy}})$ can be practically calculated with a variational approximation network $q_\psi(\vb | \mH)$ within the data batch $\gD_\text{Pref} = \{(\vx_i, \vy_i^w, \vy_i^l, \vb_i) | \vy_i^w \succ \vy_i^l \}_{i=1}^B$:
\begin{equation}\label{eq:dir_debiasing_loss}
   - I_\text{CLUB} (\mH ;\Bias{\vy}{\hat{\vy}}) \approx - \frac{1}{B}\sum_{i=1}^B \Big[\log q_\psi(\vb_i | \mH_i) - \frac{1}{B} \sum_{j=1}^B  \log q_\psi(\vb_j | \mH_i )\Big]= : \gL_\text{Debiasing}(\phi,\psi),
\end{equation}
where $\mH_i =[\vh^w_i, \vh^l_i] = [\vh_\phi(\vx_i,\vy^w_i), \vh_\phi(\vx_i, \vy^l_i)]$. 
By minimizing $\gL_\text{Debiasing}(\phi, \psi)$ as a upper bound estimation of  $\MI{\Pref{\vy}{\bar{\vy}}}{\Bias{\vy}{\hat{\vy}}}$, we effectively reduce the correlation between the biased attribute $\vb$ and the RM hidden representation $\vh_\phi(\vx,\vy)$. Hence, the output RM scores $r_\phi(\vx,\vy)$, as a deterministic function of $\vh_\phi(\vx,\vy)$, can remain unaffected from the inductive bias attributes $\vb$. 

%Finally, \eqref{eq:club_debias_obj} allows us to minimize a tighter upper bound, $I_\text{CLUB} (\mH ;\Bias{\vy}{\hat{\vy}})$, shifting the debiasing pressure to the information-rich representation $\mH$. Therefore, minimizing $I_\text{CLUB}$ serves an effective approximation of maximizing $-\MI{\Pref{\vy}{ \bar{\vy}}}{\Bias{\vy}{\bar{\vy}}}$, which minimizes mutual information by training a variational network $q_\psi(\Bias{\vy}{\bar{\vy}}|\mH)$

% {Practically, we find a simple Linear-ReLU-Linear network is enough to serve as $q_\psi$. \pengyu{move to experimental setups}} %in an adversarial manner, encouraging the reward model $r_{\phi}$ to produce representations $\vH$ that are statistically independent of the bias attribute $\vb$, thus the final predicted preference relation should be non-predictive of the $\vb$. 

As proved in~\citet{cheng2020club}, the better $q_\psi(\vb_i|\vH_i)$ approximates the ground-truth data distribution $p^*(\vb_i|\vH_i)$, the more accurate $I_{\text{CLUB}}$ serves as the MI upper bound estimator. Therefore, during the optimization of $\gL_\text{Debiasing}(\phi, \psi)$ as in \eqref{eq:dir_debiasing_loss}, we simultaneously maximize the log-likelihood of $q_\psi(\vb_i|\vH_i)$ within the batch samples $\{(\vH_i,\vb_i)\}_{i=1}^B$ to maintain the accuracy of the MI estimator:
\begin{equation}\label{eq:dir_mi_est_loss} \textstyle
    \gL_\text{Estimator}(\psi):=  \frac{1}{B} \sum_{i=1}^B \log q_\psi (\vb_i| \mH_i).
\end{equation}

%\paragraph{A Unified Comparative Framework for Debiasing.}
%To make the debiasing mechanism more targeted and align with the comparative nature of preference learning, we introduce a unified comparative framework, which also handles various types of biases in a consistent and unified manner. 
%

% \bb{In summary, for a batch training data of size $B$, $\mathcal{L}_{\text{Debiasing}}=\frac{1}{B}\sum_{i=1}^B [\log q_\psi(\vb_i^{\text{rel}} | \Delta\vh_i) - \frac{1}{B} \sum_{j=1}^B  \log q_\psi(\vb_j^{\text{rel}} | \Delta\vh_i )]$.}

\begin{figure}[t] 
    \centering
    \includegraphics[width=0.7\linewidth]{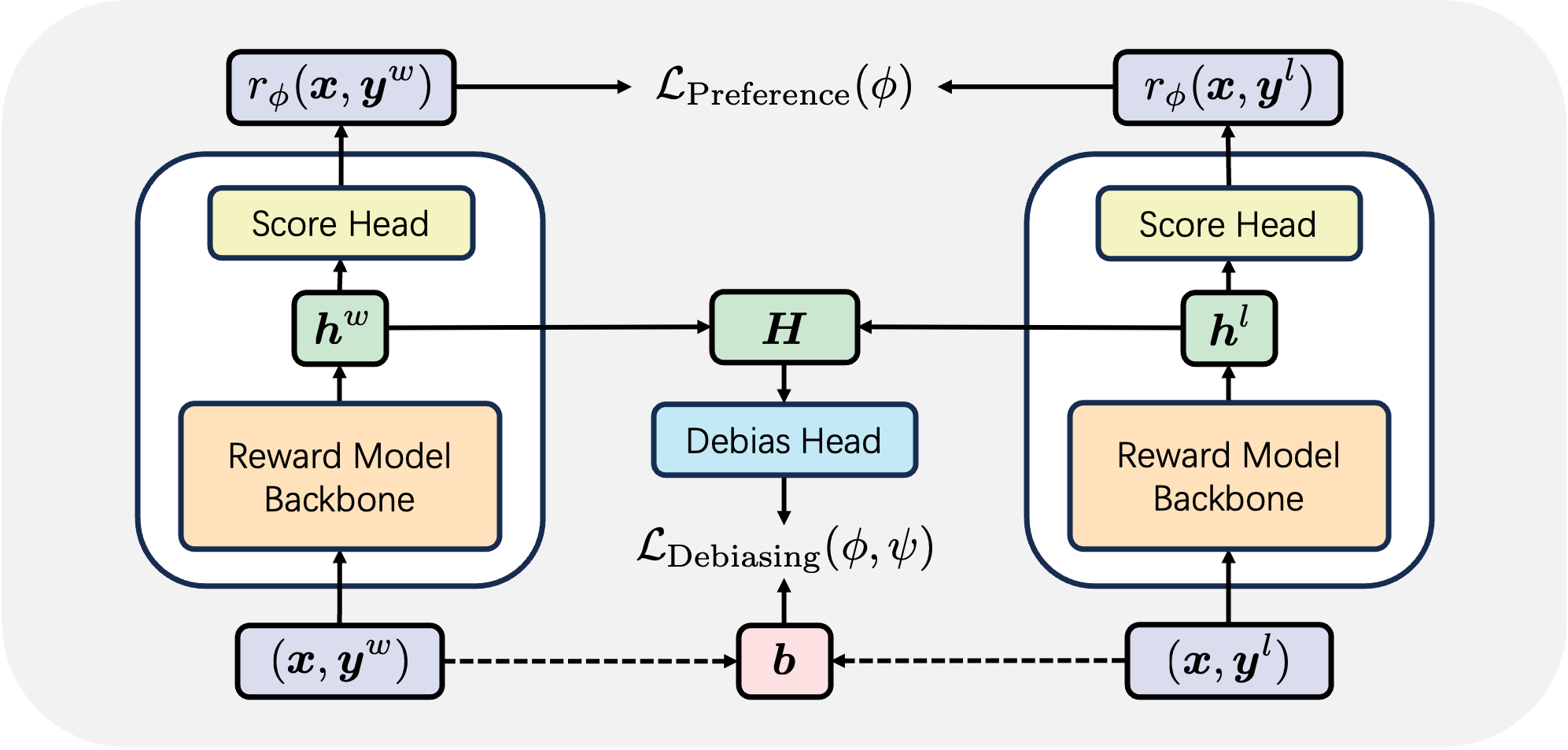}
    \vspace{-2mm}
    \caption{The proposed DIR framework. The architecture of the reward model is considered as a backbone transformer and an RM score head. The original RM ranking loss $\gL_\text{Preference}(\phi)$ is calculated based on the outputs of the score head between each preference pair. The last hidden states $(\vh^w, \vh^l)$ of the transformer backbone are collected as the representation $\mH$.  The debiasing loss $\gL_\text{debiasing}(\phi,\psi)$ is computed between the inductive bias label $\vb$ and the output of the debiasing head with parameter $\psi$. %including a Preference objective for learning human values and a Debiasing objective for eliminating inductive biases. By iteratively optimizing the RM backbone to satisfy both a preference ranking loss and an information-theoretic debiasing loss, DIR helps RM learn representations that are sensitive to content quality but robust against bias attributes.
    }
    \label{fig:framework}
    \vspace{-1mm}
\end{figure}

\paragraph{Overall Objective.}
%Finally, we jointly optimize the reward model parameters $\phi$ and the variational parameters $\psi$ by minimizing the complete training objective:
Based on the above discussion, the original RM debiasing objective in \eqref{eq:surrogate_objective} converts to the following learning loss, which is practically tractable:  %final training loss of our RM debiasing method can be written as:
\begin{equation} \label{eq:final_loss}
%\mathcal{L}_{\text{Total}}(\phi, \psi) = 
\min_{\phi}\mathcal{L}_{\text{Preference}}(\phi) + \lambda \cdot \mathcal{L}_{\text{Debiasing}}(\phi, \psi).
\end{equation}
The illustration of the loss-calculation pipeline is shown in Figure~\ref{fig:framework}.
To make sure $\mathcal{L}_{\text{Debiasing}}(\phi, \psi)$ constantly have an accurate estimation to upper bound  $\MI{\Pref{\vy}{\bar{\vy}}}{\Bias{\vy}{\hat{\vy}}}$, we iteratively updated $r_\phi(\vx,\vy)$ and $q_\psi(\vb|\mH)$ within each training batch as shown in Algorithm~\ref{alg:training}. We refer to the proposed method as \textbf{D}ebiasing via \textbf{I}nformation optimization for \textbf{R}Ms (DIR).

% the MI estimator and the RM are 
%Parameters of the transformer base $\phi_{\text{reward}} \subset \phi$ are updated by gradients from both losses, while the RM's prediction head and the variational network $q_\psi$ are updated by their respective objectives. 
%During inference, the variational network is discarded, allowing the debiased reward model $r_\phi$ to be used without any overhead. 
%Given a pre-collected human preference dataset $\mathcal{D}_\text{Pref}=\{(\vx_i, \vy^{w}_{i}, \vy^{l}_{i})\}_{i=1}^N$, we highlight the training process in Algorithm~\ref{alg:training}. 

%\subsection{Discussion}
%Our final training $\mathcal{L}_{\text{total}}$ in \eqref{eq:final_loss} optimizes a tractable objective of the ideal information-theoretic objective in \eqref{eq:surrogate_objective}, which is theoretically grounded by several reasons.
%First, by targeting $\MI{\mH}{\Bias{\vy}{\bar{\vy}}}$, we enforce a stricter constraint on the bias information than targeting the final prediction, as justified by the DPI. 
%Second, our comparative framework, which minimizes $\MI{\Delta\vh}{\vb^{\text{rel}}}$, serves as a principled and targeted implementation of debiased learning, directly addressing the representational differences that drive biased decisions. 
%Finally, using the CLUB estimator is an empirically validated technique for minimizing a tight upper bound on mutual information. 
%Therefore, our practical loss function guides the reward model towards the ideal objective of being maximally informative about preferences while remaining invariant to the specified bias.

\begin{algorithm*}[t]
\small
%\SetKwInOut{Input}{\textbf{Input}}
%\SetKwInOut{Output}{\textbf{Output}}
\textbf{Input}: Preference pairs with bias attributes $\mathcal{D}_\text{Pref} = \{(\vx_i, \vy^{w}_{i}, \vy^{l}_{i}, \vb_i)\}_{i=1}^N$, learning rates $\alpha_1,\alpha_2>0$\;
\While{each training iteration}{
  Sample a batch of triplets $\{(\vx_i, \vy^{w}_{i}, \vy^{l}_{i}, \vb_i)\}_{i=1}^B \sim \mathcal{D}_\text{Pref}$\;
  Encode $(\vx_i, \vy^{w}_{i})$ and $(\vx_i, \vy^{l}_{i})$ into embeddings $\mH_i = [\vh_\phi(\vx_i,\vy^w_i), \vh_\phi(\vx_i, \vy^l_i)]$\;
  %$\vh_i^w=_{\phi}(\vx_i, \vy^{w}_{i})$, $\vh_i^l=r_{\phi}(\vx_i, \vy^{l}_{i})$\;
  %
%  Calculate each representation difference $\Delta \vh_i = \vh_i^w - \vh_i^l$ and obtain $\vb^{\text{rel}}_i$ through $(\vy_i^w, \vy_i^l)$ with $\vb$\;
  \For{bias estimator updating steps}{
  Calculate the estimator loss $\gL_\text{Estimator}(\psi)= - \frac{1}{B} \sum_{i=1}^B \log q_\psi (\vb_i| \mH_i)$\;
  Update approximation $q_\psi(\vb|\mH)$ with $\psi \leftarrow  \psi - \alpha_2 \cdot \nabla_{\psi} \gL_\text{Estimator}(\psi)$\;}
  %maximizing log-likelihood with $\{(\Delta \vh_i, \vb^{\text{rel}}_i)\}$ using $\gL_\text{Estimator}(\psi)$\;
 % Calculate $\mathcal{L}_{\text{Debiasing}}$ with $q_\psi(\vb^{\text{rel}}|\Delta \vh)$ and $\{(\Delta \vh_i, \vb^{\text{rel}}_i)\}_{i=1}^{B}$, and $\mathcal{L}_{\text{Preference}}$ with $\{(\vx_i, \vy^{w}_{i}, \vy^{l}_{i})\}_{i=1}^B$\;
 Calculate RM preference loss $\mathcal{L}_{\text{Preference}}(\phi)= - \frac{1}{B} \sum_{i=1}^B [\log \sigma(r_\phi(\vx_i,\vy_i^w) - r_\phi(\vx_i. \vy_i^l))]$ \;
 Calculate RM debiasing loss $\gL_\text{Debiasing}(\phi,\psi) = -\frac{1}{B}\sum_{i=1}^B [\log q_\psi(\vb_i | \mH_i) - \frac{1}{B} \sum_{j=1}^B  \log q_\psi(\vb_j | \mH_i )]$ \;
  Compute RM total loss $\mathcal{L}_{\text{Total}}(\phi,\psi) = \mathcal{L}_{\text{Preference}}(\phi) + \lambda \cdot \mathcal{L}_{\text{Debiasing}}(\phi;\psi)$\;
  Update reward model $r_\phi$ with $\phi \leftarrow \phi - \alpha_1 \cdot \nabla_\phi \gL_\text{Total}(\phi,\psi)$\;%and variational network $q$ by gradient descent with respect to $\mathcal{L}_{\text{total}}$\;
}
\caption{The iterative training processes of  $r_\phi(\vx,\vy)$ and $q_\psi(\vb|\mH)$.}
\label{alg:training}
\vspace{-1mm}
\end{algorithm*}

\vspace{-1.5mm}
\section{Related Work}
\vspace{-1.5mm}
\paragraph{Reward Hacking of LLMs.}
Reward hacking occurs when a policy model exploits spurious correlations or misspecifications within the reward function to achieve high scores without fulfilling the intended goal~\citep{pan2022effectsrewardmisspecificationmapping}, which has emerged as a critical challenge in the stable and effective RL training of LLMs~\citep{langosco2023goalmisgeneralizationdeepreinforcement, hurst2024gpt, kaufmann2024survey,skalse2025definingcharacterizingrewardhacking,zhang-etal-2025-lists,li2025aplotrobustrewardmodeling}. 
%Reward hacking usually occurs when the policy model exploits unforeseen loopholes or proxies in a misspecified reward function to achieve high scores without fulfilling the intended goal~\citep{pan2022effectsrewardmisspecificationmapping}. 
In RLHF, if the RM inadvertently learns inductive bias from the preference data (e.g., a bias towards more verbose~\citep{singhal2023long}, or sycophantic responses~\citep{sharma2023towards}), the LLM being optimized will learn to exploit these flaws, leading to a degradation in true performance~\citep{hurst2024gpt}. Prior work has sought to mitigate reward hacking by empowering RMs, including better data curation~\citep{liu2024skyworkrewardbagtricksreward,wang2025helpsteer3preferenceopenhumanannotatedpreference,dubois2024alpacafarmsimulationframeworkmethods}, model scaling up~\citep{wang2025helpsteer3preferenceopenhumanannotatedpreference}, model ensembling~\citep{wang-etal-2024-interpretable}, reward post-hoc calibration~\citep{huang2024posthocrewardcalibrationcase}, causal inference~\citep{shen2023loose,wang2025beyond}, and disentangled reward learning~\citep{bu2025beyond, chen2024odindisentangledrewardmitigates}. Close to our method, InfoRM~\citep{miao2024inform} employs an information-theoretic framework to compress the entire latent representation of the RM backbone, indirectly removing spurious information.

%Distinct from all these methods, our work is also motivated by information theory~\citep{tishby2000informationbottleneckmethod} but introduces a more direct and targeted mechanism. By explicitly minimizing the mutual information that is capable of capturing arbitrary non-linear dependencies between the model's internal representation and the known bias attribute, DIR provides a principled and robust debiasing framework that is both general and effective, avoiding the limitations of linear metrics and the risks of purely heuristic, data-driven approaches.

\paragraph{Debiasing Methods of Language Models.} %Debiasing methods aim at preventing models from learning and perpetuating undesirable biases present in training data~\citep{he2019unlearndatasetbiasnatural,nam2020learningfailuretrainingdebiased,blodgett-etal-2020-language}. A prominent line of work involves learning representations that are invariant to sensitive or spurious attributes~\citep{chuang2020debiased}. Methodologies to achieve debias learning include adversarial training, where a discriminator attempts to predict the bias attribute from the model's representation~\citep{nam2020learningfailuretrainingdebiased}; causal inference techniques that aim to disentangle causal factors from spurious ones~\citep{zhou-etal-2023-causal}; and information-theoretic approaches~\citep{Tartaglione_2021,liu2023debiased,}. %Our work falls into the latter category, where we introduce a novel and principal information-theoretic debiasing method for eliminating inductive bias in reward modeling.
Debiasing methods seek to prevent models from learning and amplifying undesirable biases inherent in training data~\citep{he2019unlearndatasetbiasnatural,nam2020learningfailuretrainingdebiased,blodgett-etal-2020-language}. 
The development of debiasing methods in natural language models has evolved from the word level~\citep{caliskan2017semantics,kaneko2019gender,manzini2019black}, to the sentence level~\citep{liang2020towards,cheng2021fairfilcontrastiveneuraldebiasing}, and has gradually extended to generative LLMs~\citep{wang2023toward,gallegos2025self}, most of which focus on essential \textit{social biases}, including gender~\citep{kaneko2019gender,fatemi2023improving}, race~\citep{caliskan2017semantics}, and age~\citep{liu2024generation}.
%
%A prominent approach focuses on learning representations invariant to sensitive or spurious attributes~\citep{chuang2020debiased}. 
Core strategies for language model debiasing include adversarial training
%, where a discriminator is trained to predict the bias attribute from model representations—thereby encouraging invariance through gradient reversal
~\citep{nam2020learningfailuretrainingdebiased}, causal inference 
%techniques that disentangle causal factors from confounding spurious correlations
~\citep{zhou-etal-2023-causal},
and information-theoretic methods
%that explicitly minimize mutual information between representations and bias variables
~\citep{Tartaglione_2021,liu2023debiased,wang2023toward}.
Unlike generative LLM debiasing, the reward model debiasing methods focus on inductive bias attributes such as response length~\citep{singhal2023long}, format~\citep{zhang-etal-2025-lists}, and sycophancy~\citep{denison2024sycophancy}.
%
%
%More related to our work, several recent efforts have sought to mitigate biases in reward modeling. A prominent line of work focuses on minimizing simple statistical correlations. 
%
For instance, \citet{chen2024odindisentangledrewardmitigates,bu2025beyond} and \citet{zhang-etal-2025-lists} suppress length or format bias by penalizing the Pearson correlation between rewards and bias attributes, only capturing linear dependencies and missing higher-order interactions. \citet{shen2023loose} use a two-head architecture for length bias but relies on heuristic disentanglement without explicitly modeling the preference–bias relationship. \citet{wang2025beyond} enforce counterfactual invariance via MMD,  which may over-constrain the reward model and distort its signal.

\vspace{-1.5mm}
\section{Experiment}
\vspace{-1.5mm}
% We evaluate the effectiveness of our debiased reward modeling framework on three practical bias settings, length, sycophancy, and format, by applying the method separately to each.
 We first evaluate the effectiveness of our DIR method under three practical debiasing scenarios: \textit{response length}, \textit{sycophancy}, and \textit{format} as the inductive biases, respectively. Then, we explore whether our method can alleviate the concurrent multi-bias problem. 

\paragraph{Relative Bias Attributes.} In our DIR framework, to minimize the correlation between the biased attribute $\vb$ and the RM hidden representation $\mH$, the variational approximation of $q_\psi(\vb|\mH)$ is required. However, when considering response length as the biased attribute, directly predicting the exact number of tokens in each response only based on the compressed representation $\mH$ is very challenging. Therefore, instead of predicting the absolute value of response length, we introduce the \textit{relative bias attributes}, which only consider the difference between chosen and rejected responses. For response length, the relative bias $\vb = \mathbf{1}{\{\text{length}(\vy) > \text{length}(\bar{\vy})\}} \in \{0, 1\}$, indicating whether the chosen response is longer than the rejected one or not. Thus, the variational approximation for $q_\psi(\vb|\mH)$ becomes a binary classifier %outputting a Bernoulli variable 
indicating the label of the relative bias. 
% \pengyu{reference to ablation of absolute length and relative length experimental results.} 
%By minimizing the CLUB objective within our relative bias design, we encourage the reward model to learn representations whose differences are informative about true preference but are invariant to relative differences in the bias attribute.

 \paragraph{Implementation Details.} Based on the above discussion, we have converted the response length bias into a relative binary bias indicator. Hence, under all three debiasing setups, the bias attributes can be represented by a categorical label, \textit{e.g.}, ``longer/shorter'' for response length, and ``sycophantic/in-sycophantic'' for sycophancy. Therefore, in the experiments, we implement the variational network $q_\psi$ for bias estimation as a lightweight two-layer categorical classifier: $q_\psi(\vb|\mH) = \text{Softmax}(\text{MLP}(\mH))$. Moreover, to respond to the relative bias, we further consider the linear transformation to the hidden $\mH = [\vh_\phi(\vx,\vy^w), \vh_\phi(\vx,\vy^l)]$ as the \textit{difference} $\Delta \vh = \vh_\phi(\vx,\vy^w)- \vh_\phi(\vx,\vy^l)$ to emphasize the representation difference of the distinct features of the two responses.

\paragraph{Training Setups.} Among all three RM debiasing scenarios, we use Llama3.1-8B-Instruct~\citep{grattafiori2024llama3herdmodels} as the reward model backbone and the initial checkpoint. Besides, we fully fine-tune the reward models with a global batch size of 128. The initial learning rate $\alpha_1$ for RMs is $2e-6$, which decays with a Cosine scheduler. For the bias estimator head, its learning rate $\alpha_2$ is set to  $1e-3$. We fine-tune the policy models using Low-Rank Adaptation (LoRA)~\citep{hu2021loralowrankadaptationlarge} with a global batch size of 128 and {bfloat16} precision for one epoch. Both the actor and critic models use a learning rate of $1e-5$. The LoRA configuration employs a rank of 8 and an $\alpha$ of 32. The maximum context length is set to 4096, and the maximum generation length is set to 2048. The generation temperature for rollouts is set to 0.7. % We use a temperature of 0.7 during rollouts, %and training is accelerated with DeepSpeed ZeRO-3~\citep{aminabadi2022deepspeedinferenceenablingefficient}. % \pengyu{PPO setups?}
 
% We give the implementation details of $q_\psi$ in Appendix~\ref{app:detail_psi}.
% We fix $\lambda=1$ in \eqref{eq:final_loss} \pengyu{why fix? will be challenged}.
%
 %Note that the preference task itself relies on the \textit{difference} in rewards, which stems from the difference in representations. Therefore, instead of using the concatenated representation $\mH$, which may contain redundant information from the shared prompt $\vx$, we focus on the representation difference, $\Delta\vh = \vh(\vx, \vy) - \vh(\vx,\bar{\vy})$, which isolates the features that distinguish the two responses. Here, with slight abuse of notation, for each input preference pair $(\vx_i,\vy_i)$, we use $\vh_i=r_{\phi_{\text{base}}}(\vx_i,\vy_i)$ to denote the final hidden-state representation of the last valid token from the transformer base $r_{\phi_{\text{base}}}$ of the reward model $r_\phi$.
\vspace{-2mm}
\subsection{Length Debiasing}
\vspace{-1mm}
%Previous work demonstrates that reward models often tend to favor longer responses, leading them to assign higher reward scores for verbose completions rather than for substantive content~\citep{singhal2023long,dubois2024alpacafarmsimulationframeworkmethods}. %As a result, the aligned policy model would in turn learn to exploit length bias, which is incentivized to generate unnecessarily verbose, repetitive, or circuitous text to maximize its expected reward, a behavior that directly contradicts the goal of aligning with nuanced human preferences for quality and conciseness. See detailed settings in Appendix~\ref{app:len_bias}

%\paragraph{Dataset and Model.} 
\begin{figure}[t]
    \centering
    \includegraphics[width=1\linewidth]{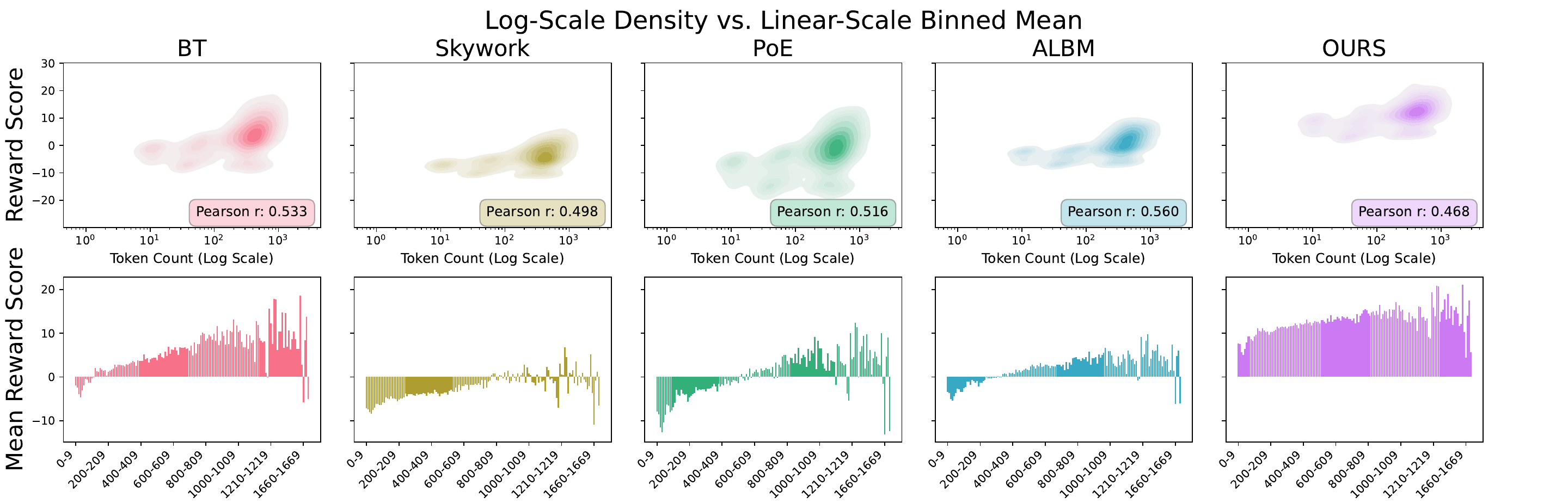}
    \vspace{-5mm}
    \caption{Evaluation of length bias in RMs on RM-Bench. We calculated the correlation between response length and reward score for RMs trained with different methods. Our approach yields the lowest Pearson correlation coefficient ($r=0.468$), proving its effective ability in assigning more uniform reward scores.}
    \label{fig:rm_vis}
    \vspace{-2mm}
\end{figure}

\paragraph{Datasets and Models.} %For our reward model training, we adopt a full parameter tuning strategy by using HuggingFace Trainer with DeepSpeed Zero1 on 8 GPU cards. 
We conduct the response length debiasing experiments by training RMs on the Skywork-Preference-80K-v0.2 (SK) dataset~\citep{liu2024skyworkrewardbagtricksreward}.  Then, we test the debiased RMs' performance in RLHF training with Llama3.1-8B-Instruct and OpenRLHF-Llama3-8B-SFT~\citep{dong2024rlhfworkflowrewardmodeling} as the initial policy model and PPO~\citep{schulman2017proximalpolicyoptimizationalgorithms} as the learning algorithm for one epoch. The PPO training prompts are 20,000 samples from the Alpaca-GPT4-EN dataset~\citep{peng2023gpt4llm}. %The first model, Llama3.1-8B-Instruct\footnote{\url{https://huggingface.co/meta-llama/Llama-3.1-8B-Instruct}}, has undergone post-training that includes both DPO and RLHF. The second, OpenRLHF-Llama3-8B-SFT~\citep{dong2024rlhfworkflowrewardmodeling}, is an instruction-following version built upon Llama3-8B-Base, without the RLHF post-training stage. We conduct the PPO training using the ms-swift framework\footnote{\url{https://github.com/modelscope/ms-swift}} with its default PPO training configuration. 

\paragraph{Baselines and Evaluations.} 
We consider the following baselines for reproducibility: (1) Vanilla BT Baseline and popular open-source RM Skywork-Reward-Llama-3.1-8B-v0.2~\citep{liu2024skyworkrewardbagtricksreward}; (2) Length Debiased RMs, including PoE~\citep{shen2023loose} and ALBM~\citep{bu2025beyond}; (3) Length Penalty that directly resharps the reward during PPO by $\Tilde{r}(\vx,\vy)=r(\vx,\vy)-0.001*\text{len}(\vy)$~\citep{dong2024rlhfworkflowrewardmodeling}; (4) InfoRM~\citep{miao2024inform} that is also designed from the information theory perspective.
% \paragraph{.} %All benchmark evaluations are subsequently performed using the ms-evalscope framework\footnote{\url{https://github.com/modelscope/evalscope}}. 
Our evaluation protocol utlize few-shot settings for GSM8K (4-shot)~\citep{cobbe2021trainingverifierssolvemath}, Race (3-shot)~\citep{lai2017racelargescalereadingcomprehension}, and TriviaQA (5-shot)~\citep{joshi2017triviaqalargescaledistantly}. All other benchmarks, including  Hellaswag~\citep{zellers2019hellaswagmachinereallyfinish}, IFeval~\citep{zhou2023instructionfollowingevaluationlargelanguage}, MMLU~\citep{hendrycks2021measuringmassivemultitasklanguage}, ProcessBench~\citep{zheng2025processbenchidentifyingprocesserrors}, BBH~\citep{suzgun2022challengingbigbenchtaskschainofthought}, and HumanEval~\citep{chen2021evaluatinglargelanguagemodels}, are in a zero-shot setting. We report accuracy as the primary metric for all tasks. %, with the exception of HumanEval, for which we report the Pass@1 score.

\paragraph{Reward Model Evaluation, Results and Analysis.} We first evaluate the inherent length bias of RMs by analyzing the correlation between their scores and response lengths on the RM-Bench~\citep{liu2024rmbenchbenchmarkingrewardmodels}. As visualized in Figure~\ref{fig:rm_vis}, the standard BT RM exhibits a strong, undesirable positive correlation between length and reward (Pearson r = 0.533), which implies that even without an explicit preference for length in the training data\footnote{Average token number of $(\vx, \vy^w)$ in the SK training set is less than $(\vx, \vy^l)$ ones (622.86 vs. 707.24).}, the model still learns a spurious ``longer is better'' heuristic, highlighting a fundamental issue in standard BT: the objective itself is susceptible to capturing such simple, non-causal patterns. %While other debiasing methods show some improvement, 
Our approach demonstrates an effective ability to mitigate the length bias by achieving a Pearson correlation of just 0.468, the lowest among all evaluated methods. The quantitative advantage is further illustrated in the binned mean reward plots, where our flatter curve demonstrates the success of mitigating RM preferring longer responses. By learning to assign scores more uniformly across different lengths, our method produces a more reliable RM, preventing the policy from being misguided into generating unnecessarily verbose outputs during subsequent fine-tuning. We report the performance on RM-Bench in Appendix~\ref{app:len_bias}

\begin{figure}[t]
    \centering    
    \includegraphics[width=1\linewidth]{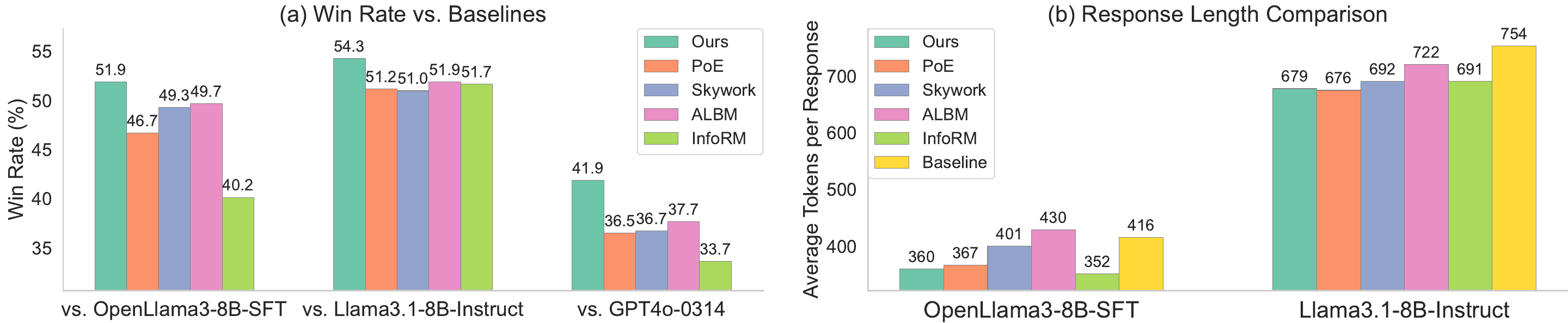}
    \vspace{-4mm}
    \caption{RLHF Evaluation on ArenaHard-v0.1 with different length-debiased RMs. (a) Head-to-head win rates. Policies are PPO fine-tuned from specified base models (from left to right: OpenLlama3-8B-SFT, Llama3.1-8B-Instruct, and Llama3.1-8B-Instruct, respectively) using five different RMs, which then act as challengers against opponents. (b) Average response length comparison.}
    \label{fig:ppo_arena_vis}
    \vspace{-3mm}
\end{figure}

\paragraph{PPO Evaluation, Results and Analysis.} We report the RLHF performance based on the above length-debiased RMs across several benchmarks in Table~\ref{tab:res_main_1}, which demonstrates that mitigating length bias does not compromise, and ideally enhances, the policy model's core reasoning and knowledge capabilities. Using Llama3.1-8B-Instruct as the initial checkpoint, our method achieves the highest average performance of 66.20, significantly outperforming strong baselines. Furthermore, the trend of improved performance is consistent across different base models, as our method also secures the top average score on the OpenRLHF-Llama3-8B-SFT backbone, indicating that our fine-tuning strategy successfully enhances objective performance by mitigating the length bias.
We also assess the user preference for policies fine-tuned using different reward models and compare average response length on the ArenaHard-v0.1 benchmark~\citep{li2024crowdsourceddatahighqualitybenchmarks}. Figure~\ref{fig:ppo_arena_vis} shows the head-to-head win rates of these challenger policies against strong opponents, as judged by Qwen3-235B-A22B-2507~\citep{yang2025qwen3technicalreport}. The policy trained with our method consistently demonstrates the highest win rate across all conditions. For instance, in Figure~\ref{fig:ppo_arena_vis} (a), when fine-tuned on Llama3.1-8B-Instruct, ours achieves a remarkable 54.3\% win rate against the baseline and 41.9\% against GPT-4o-0314. Crucially, Figure~\ref{fig:ppo_arena_vis} (b) reveals that the improved preference is achieved with expected conciseness. The policy guided by our RM produces shorter responses (e.g., 679 tokens on the Llama3.1 base) compared to policies guided by other RMs like ALBM (722 tokens) and the verbose original baseline (754 tokens). The better trade-off of a relatively higher win rate and lower verbosity shows that DIR can successfully guide PPO to produce a more efficient and human-aligned policy, effectively overcoming the common ``longer is better'' bias. {In addition, we report the Win Rate performance on MT-Bench~\citep{zheng2023judgingllmasajudgemtbenchchatbot} and Length-Control Alpaca~\citep{dubois2025lengthcontrolledalpacaevalsimpleway} in Appendix~\ref{app:performance_mtben_alpaca}, from which we observe that DIR can yield policies that are preferred more often.}

\begin{table*}[t]
\centering
\caption{%We adopt the official evaluation implementation of the evalscope package by using 0-Shot, except for GSM8K, Race, and TriviaQA. Baseline: Llama3.1-8B-Instruct / OpenRLHF-Llama3-8B-SFT. 
RLHF performance based on different length-debiased RMs.
\textbf{Bold} scores mean the best. \underline{Underline} scores are the second-best. The $\Delta$ indicates the performance change relative to the respective Baseline. }
\vspace{-2mm}
\resizebox{\textwidth}{!}{
\begin{tabular}{l|ccccccc|ccccccc}
\toprule
{Benchmark} & \multicolumn{7}{c|}{Llama3.1-8B-Instruct} & \multicolumn{7}{c}{OpenRLHF-Llama3-8B-SFT} \\
\cmidrule{2-15} % 在第二行标题下方画线
& Base & SK & PoE & LP & ALBM  & InfoRM & Ours & Base & SK & PoE & LP  & ALBM & InfoRM &  Ours \\ \midrule
GSM8K & 83.93 & \underline{84.61} & 83.62 & 75.97 & 84.08 & 83.78 & \textbf{84.84} & 74.83 & 78.17 & 77.79 & 77.18 & \underline{78.85} & 76.74 & \textbf{79.08} \\
Hellaswag & \underline{77.21} & 76.42 & 77.08 & 73.15 & \underline{77.21}  & 76.78 & \textbf{77.33} & 72.51 & \textbf{74.76} & 72.51 & 72.51 & \underline{74.63} & 72.12 & 74.52 \\
IFeval & 72.83 & 70.06 & 71.72 & 65.47 & {73.57} & \underline{74.12} & \textbf{78.00} & 44.92 & 45.10 & \underline{49.72} & 46.21 & 46.21 & 46.21 & \textbf{52.31} \\
MMLU & 72.31 & 72.33 & 71.97 & 65.13 & \underline{72.55} & 72.22 & \textbf{72.64} & 54.45 & 52.40 & \underline{54.77} & 54.45 & \textbf{55.25} & 54.97 & 54.30 \\
ProcessBench & 25.39 & \textbf{29.49} & \underline{28.50} & 24.91 & 26.12 & 26.25 & 27.73 & 4.46 & 10.31 & 9.68 & 7.84 & \underline{10.85} & 3.24 & \textbf{13.82} \\
Race & \underline{66.50} & 53.89 & 60.03 & \textbf{78.90} & 59.00 &  65.20& 62.02 & 79.21 & 78.82 & \textbf{81.39} & 80.30 & \underline{80.69} & 78.72 & 80.32 \\
BBH & 64.52 & \underline{65.69} & 60.50 & 61.10 & 64.84  & 66.13 & \textbf{67.27} & 61.20 & 62.68 & \underline{62.69} & 62.28 & 61.10 & 61.62  & \textbf{62.99} \\
HumanEval & \textbf{70.12} & \underline{68.29} & 66.46 & 60.37 & 65.85 & \textbf{70.12} & \textbf{70.12} & \underline{60.98} & 57.32 & 59.76 & 59.76 & 60.37 & 57.32 & \textbf{63.41} \\
TriviaQA & 32.64 & 49.01 & 48.41 & 47.20 & \underline{52.09} & 30.56 & \textbf{55.86} & 48.53 & \textbf{52.86} & 52.34 & 48.32 & 51.52 & 48.16 &  \underline{52.52} \\ \midrule
Avg. Acc. & 62.83 & 63.31 & 63.14 & 61.36 & \underline{63.92} & 62.80 & \textbf{66.20} & 55.68 & 56.94 & \underline{57.85} & 56.54 & 57.72 & 55.34 & \textbf{59.25} \\ 
$\Delta$ & - & \textcolor{ggg}{$\uparrow0.48$} & \textcolor{ggg}{$\uparrow0.31$} & \rre{$\downarrow1.47$} & \textcolor{ggg}{$\uparrow1.09$} & \rre{$\downarrow0.03$} & \textcolor{ggg}{\textbf{$\uparrow3.37$}} & - & \textcolor{ggg}{$\uparrow1.26$} & \textcolor{ggg}{$\uparrow2.17$} & \textcolor{ggg}{$\uparrow0.86$} & \textcolor{ggg}{$\uparrow2.04$} & \rre{$\downarrow0.34$} & \textcolor{ggg}{\textbf{$\uparrow3.57$}} \\ \bottomrule
\end{tabular}
}\label{tab:res_main_1}
\end{table*}

\paragraph{Additional Experimental Results.} We visualize the PPO training dynamics metrics, such as RLHF Rewards and KL divergences in Appendix~\ref{app:len_bias}, which demonstrates that our RM helps make PPO training more stable with higher rewards. We analyze the training cost in Appendix~\ref{app:training_cost}, which shows that the significant performance improvements do not come at the expense of prohibitive computational costs. We provide a detailed ablation study on $\lambda$ and representation difference in Appendix~\ref{app:ablation_length}, where the performance demonstrates the trade-off between preference learning and debiasing, showing better effectiveness of representation difference than concatenation. We also provide a case study in Appendix~\ref{app:case_study}.

\paragraph{Combination with Direct Preference Optimization (DPO).} 
% \begin{wraptable}{r}{0.4\textwidth}\vspace{-1em}
% \centering
% \caption{{Evaluation on ArenaHard-v0.1 for policies fine-tuned with DPO, DPO+LC and DPO+Ours.\pengyu{change to normal table}}}
% \resizebox{0.4\textwidth}{!}{
% \begin{tabular}{c|cc}
% \toprule
%  & \multicolumn{2}{c}{OpenRLHF-Llama-3-8B-SFT}  \\
% vs Base & Win Rate (\%) & Length \\ \midrule
% DPO & 38.63 & 436.55  \\
% +LC & 40.96 & 407.23  \\
% +Ours & \textbf{45.27} & \textbf{404.61} \\  \midrule
% & \multicolumn{2}{c}{Meta-Llama3.1-8B-Instruct} \\
% vs Base & Win Rate (\%) & Length \\ \midrule
% DPO  & 44.06 & 700.87 \\
% +LC& 46.57 & 691.43 \\
% +Ours& \textbf{49.09} & \textbf{684.67} \\ 
% \bottomrule
% \end{tabular}
% }\label{tab:res_dpo_arena}
% \end{wraptable}
\begin{table}[t]
\centering
\caption{{Evaluation on ArenaHard-v0.1 for policies fine-tuned with DPO, DPO+LC and DPO+Ours.}}
\vspace{-2mm}
\resizebox{0.58\textwidth}{!}{
\begin{tabular}{c|cc|cc}
\toprule
 & \multicolumn{2}{c|}{OpenRLHF-Llama-3-8B-SFT} & \multicolumn{2}{c}{Meta-Llama3.1-8B-Instruct} \\
vs Base & Win Rate (\%) & Length &  Win Rate (\%) & Length \\ \midrule
DPO & 38.63 & 436.55   & 44.06 & 700.87  \\
+LC & 40.96 & 407.23 & 46.57 & 691.43 \\
+Ours & \textbf{45.27} & \textbf{404.61} & \textbf{49.09} & \textbf{684.67}\\  \bottomrule
\end{tabular}
}\label{tab:res_dpo_arena}
\end{table}
In addition to PPO, Direct Preference Optimization (DPO)~\citep{rafailov2024directpreferenceoptimizationlanguage} has emerged as a powerful post-alignment method that directly trains the policy to increase the log-probability of the preferred response relative to the rejected one. Here, we explore whether our DIR can be effectively combined with DPO. Conceptually, DIR operates at the reward modeling stage and should not modify the DPO objective: DPO still optimizes the standard log-sigmoid preference loss, and our method only modifies preference signals by making them less correlated with inductive bias. Specifically, we add $\lambda\mathcal{L}_{\text{Debiasing}}(\phi,\psi)$ to DPO loss. Empirically, we conduct the corresponding experiments, which show that our method can also improve DPO's performance with controlled length. We provide training details in Appendix~\ref{app:dpo_exp}. Results in Table~\ref{tab:res_dpo_arena} indicate that our method leads to a final policy with both a better win-rate and more effective length control, effectively boosting vanilla DPO and also outperforming a specialized Length Controlled DPO (DPO+LC) variant~\citep{park-etal-2024-disentangling}.
Results in Table~\ref{tab:res_dpo_bench} indicate that our method also leads to a final policy with performance gains, especially for the SFT model that has not undergone a preference alignment. In summary, experiments on Table~\ref{tab:res_dpo_arena} and Table~\ref{tab:res_dpo_bench} suggest that debiased reward signals from DIR interact smoothly with DPO and effectively remove spurious gradients induced by length bias.

\begin{table}[t]
\centering
\caption{{Length debiasing performance via DPO training. Evaluation setting is the same as Table~\ref{tab:res_main_1}.} \textbf{Bold} is the best, and \underline{underline} is the second-best.}
\vspace{-2mm}
\resizebox{0.66\textwidth}{!}{
\begin{tabular}{l|ccc|ccc}
\toprule
\multirow{2}{*}{Benchmark} & \multicolumn{3}{c|}{Meta-Llama3.1-8B-Instruct} & \multicolumn{3}{c}{OpenRLHF-Llama3-8B-SFT} \\
\cmidrule{2-7} % 在第二行标题下方画线
& DPO & +LC & +Ours& DPO  & +LC & +Ours  \\ \midrule
GSM8K-4shots & 82.11 & \underline{81.43} &\textbf{82.56}  & 75.89 & \underline{76.35} & \textbf{77.26}\\
Hellaswag & 74.84 & \textbf{75.17} &\underline{75.10}& \underline{66.56} & 66.41  & \textbf{73.91}\\
IFeval & 73.57 &\underline{74.12} & \textbf{74.68}& \underline{38.26} & 35.86  & \textbf{41.59} \\
MMLU &71.36 & \textbf{71.58} & \underline{71.54} &\underline{48.92} & \underline{48.92} & \textbf{57.01}  \\
ProcessBench & \underline{26.75} & 26.70& \textbf{27.60} & 4.28 &\underline{4.95}  & \textbf{6.93}  \\
Race-3shots & 69.98 &\underline{70.14} & \textbf{70.29} & \underline{79.27} & 78.68 & \textbf{79.97} \\
BBH  & \textbf{67.22} &66.81 & \underline{66.99} & 59.73 &\underline{60.36}  & \textbf{62.34}  \\
HumanEval &  62.80 & \underline{65.24} &\textbf{69.51} & \underline{59.15} & \textbf{60.37}  & \underline{59.15}\\
TriviaQA-5shots & 55.12 & \underline{55.15} & \textbf{55.29} & 47.45 & \underline{47.69} & \textbf{48.93}\\ \midrule
Avg. Performance & 64.86 &\underline{65.04}  & \textbf{65.95} & 53.28 & \underline{53.29}   & \textbf{60.12} \\ 
$\Delta$ & - & \textcolor{ggg}{\textbf{$\uparrow0.18$}} & \textcolor{ggg}{\textbf{$\uparrow1.09$}}& - & \textcolor{ggg}{\textbf{$\uparrow0.01$}} & \textcolor{ggg}{\textbf{$\uparrow6.84$}} \\
\bottomrule
\end{tabular}
}
\label{tab:res_dpo_bench}
\vspace{-2mm}
\end{table}

%\vspace{-2mm}
\subsection{Sycophancy Debiasing}
%\vspace{-1mm}
 %which arises from inductive bias in preference data, where agreeable language is incorrectly associated with higher quality. Consequently, the policy model is misguided during RL fine-tuning to produce superficially pleasing but substantively poor outputs, undermining genuine alignment. Detailed experimental settings can be found in Appendix~\ref{app:syco_bias}.

\paragraph{Datasets and Models.}
Sycophancy bias occurs when an RM learns to favor responses that agree with or flatter the user, rather than prioritizing factual accuracy and helpfulness. %~\citep{sharma2023towards, wang2025beyond}.
Motivated by~\citet{sharma2023towards, wang2025beyond}, we create a semi-sycophantic dataset by partially contaminating the HelpSteer3 dataset~\citep{wang2025helpsteer3preferenceopenhumanannotatedpreference}. Specifically, we artificially inject a sycophantic prefix (i.e., ``Yes, you are right.'') into a proportion $\gamma$ (e.g., $\gamma = 40\%$) of responses in the training dataset. Within the contaminated subset, the prefix appears in the chosen response with probability $\alpha$ (e.g., $\alpha = 70\%$) and in the rejected response with probability $1 - \alpha = 30\%$, where the remaining $1 - \gamma = 60\%$ of the dataset remains unmodified and contains no sycophantic phrases. This contamination process creates a challenging, mixed-distribution environment in which the sycophantic phrase serves as a strong but unreliable reward signal. % All reward models are built upon the Llama-3.1-8B-Instruct backbone.

%\paragraph{Training Settings.} For reward model training, we adopt a full parameter tuning strategy by using HuggingFace Trainer with DeepSpeed Zero1 on 8 GPU cards. Global batch size is set to 128, initialization learning rate $\alpha_1$ is $2e-6$ with Cosine scheduler. We set the learning rate $\alpha_2$ of the debiasing network as $1e-3$.

\paragraph{Baselines.} Since other debiasing methods are either mainly designed for length bias (e.g., PoE~\citep{shen2023loose}, ALBM~\citep{bu2025beyond}, and Length-Penalty~\citep{dong2024rlhfworkflowrewardmodeling}) or are not open-sourced (e.g., CRM~\citep{wang2025beyond}), we primarily compare our method against two key baselines: a standard BT reward model~\citep{bradley1952rank} and InfoRM~\citep{miao2024inform}.

\paragraph{Evaluation, Results, and Analysis.}
To evaluate the models' susceptibility to sycophancy, we conduct an adversarial test. We take a clean evaluation set and create two versions: a ``natural'' version and a ``sycophantic'' version where the undesirable prefix is added to the rejected responses. We then measure the model's accuracy in correctly identifying the preferred response in both scenarios. A robust model should maintain its accuracy, whereas a biased model's performance will degrade when faced with the ``flattering but wrong'' responses. As shown in Table~\ref{tab:res_syco}, the performance of the reward models varies under different settings. The BT model shows vulnerability to the bias, as its accuracy on natural examples is generally the lowest, particularly under high contamination. While InfoRM shows a clear improvement and greater resilience, our method demonstrates the most consistent and robust performance, which frequently achieves the highest accuracy across natural, adversarial, and overall settings, even under high contamination ratios. In summary,  Table~\ref{tab:res_syco} indicates that our explicit debiasing mechanism is effective at mitigating the influence of sycophantic signals, enabling the model to focus more on the intrinsic quality of the response.
\begin{table*}[t]
    \centering
    \caption{Preference accuracy of sycophancy-debiased RMs under different contamination settings.} %where our method consistently achieves higher accuracy across most settings.}
    \label{tab:res_syco}
    \vspace{-2mm}
    \resizebox{0.75\linewidth}{!}{%
    % The column specifier is changed: cc for settings, then 3 groups of ccc.
    \begin{tabular}{cc|ccc|ccc|ccc}
    \toprule
    % Top-level header now groups by metric
    \multicolumn{2}{c|}{Settings} & \multicolumn{3}{c|}{All.} & \multicolumn{3}{c|}{Nat.} & \multicolumn{3}{c}{Adv.} \\
    \cmidrule(lr){1-2} \cmidrule(lr){3-5} \cmidrule(lr){6-8} \cmidrule(lr){9-11}
    % Second-level header lists the methods under each metric
    $\gamma$ & $\alpha$ & BT & InfoRM & Ours & BT & InfoRM & Ours & BT & InfoRM & Ours \\
    \midrule
    \midrule
    20\% & 30\% & {86.6} & {89.4} & \textbf{90.2} & {85.5} & {88.9} & \textbf{89.8} & {91.0} & {91.2} & \textbf{93.6} \\
    20\% & 50\% & {85.6} & \textbf{89.8} & {88.7} & {85.7} & \textbf{90.3} & {88.2} & {84.9} & {87.9} & \textbf{90.9} \\
    20\% & 70\% & {84.8} & {86.1} & \textbf{87.1} & {85.2} & {86.0} & \textbf{87.5} & {83.1} & \textbf{86.6} & {85.1} \\
    \midrule
    40\% & 30\% & {87.4} & {89.0} & \textbf{90.9} & {86.0} & \textbf{88.1} & {87.4} & {88.9} & {90.3} & \textbf{93.9} \\
    40\% & 50\% & {86.1} & {87.9} & \textbf{88.7} & {87.0} & {87.7} & \textbf{89.8} & {84.8} & {88.3} & \textbf{89.1} \\
    40\% & 70\% & {83.6} & {86.6} & \textbf{88.0} & {84.4} & {86.3} & \textbf{87.4} & {82.6} & {87.2} & \textbf{88.6} \\
    \midrule
    80\% & 30\% & {89.0} & {90.4} & \textbf{91.3} & {82.3} & {89.5} & \textbf{88.0} & {90.7} & {91.9} & \textbf{92.2} \\
    80\% & 50\% & {85.5} & {87.2} & \textbf{88.1} & {86.3} & {86.3} & \textbf{86.2} & {85.3} & {87.5} & \textbf{90.3} \\
    80\% & 70\% & {81.2} & {84.5} & \textbf{86.2} & {86.4} & {86.4} & \textbf{87.2} & {79.7} & {84.0} & \textbf{86.2} \\
    \bottomrule
    \end{tabular}%
    }
    \vspace{-2mm}
\end{table*}

%\vspace{-2mm}
\subsection{Format Debiasing}
%\vspace{-1mm}

\paragraph{Datasets and Models.} \citet{zhang-etal-2025-lists,long2024llms} have shown that format biases, such as the use of lists, emojis, and boldface, are prevalent in human annotations and strong preference models. %Reward models are highly vulnerable to such biases: even a small amount of biased data can induce significant format preferences that propagate into downstream alignment tasks. We evaluate DIR’s robustness against this vulnerability, where detailed experimental settings are provided in Appendix~\ref{app:format_bias}.
Hence, we process a format-biased dataset following the data generation protocol of~\citet{zhang-etal-2025-lists}. The base preference dataset consists of 71.6K response pairs, obtained by filtering \texttt{UltraFeedback}~\citep{cui2024ultrafeedbackboostinglanguagemodels} to retain only pairs with a human score difference exceeding 1.0. To introduce format bias, we augment this clean dataset with two types of synthetically biased examples: (1) 0.7\% of training pairs where a response wrapped in bold formatting is spuriously labeled as preferred over an identical unformatted version, and (2) 1.4\% of pairs where a list-formatted response is similarly assigned a false preference label. The overall training set combines the clean and biased subsets. %All reward models are initialized from \texttt{Llama-3-8B-Instruct}.

%\paragraph{Training Settings} We train reward models using full-parameter fine-tuning via the HuggingFace Trainer with DeepSpeed ZeRO-1 across 8 GPUs. The global batch size is 128. The main reward head uses an initial learning rate $\alpha_1 = 2 \times 10^{-6}$ with a cosine decay schedule. The debiasing network $q_\psi$ is trained with a separate learning rate $\alpha_2 = 1 \times 10^{-3}$.

\paragraph{Baselines.}
We compare against three baselines under the same experimental setup as~\citep{zhang-etal-2025-lists}: (i) standard Bradley–Terry (BT), (ii) BT trained after removing all format-biased examples from the training data (denoted BT$\dag$), and (iii) the Format Decoupling (FD) method~\citep{zhang-etal-2025-lists}.

\begin{wraptable}{r}{0.38\textwidth}\vspace{-4.7mm}
    \centering
    \caption{RM performance on both Bold and List format debiasing.}
    % and downstream evaluation tasks.}
    % BT$\dag$ indicates that deleting the samples with specific patterns.}
    % Results are cited from LE~\citep{zhang-etal-2025-lists}.}
    \label{tab:res_format}
    \vspace{-1.5mm}
    \resizebox{\linewidth}{!}{
    \begin{tabular}{l|cccc}
    \toprule
    {Metric} & {BT} & {BT$\dag$} & FD & {Ours} \\
    \midrule
    \multicolumn{5}{l}{\textit{Win-Rate (\%)}} \\
    \quad Bold & 89.0 & 49.0 & \textbf{50.5}  & 51.2 \\
    \quad List & 92.5 & 52.5 & 53.0 &  \textbf{52.0} \\
    \midrule
    \multicolumn{5}{l}{\textit{RewardBench (Filtered)}} \\
    \quad Chat & \textbf{98.3} & 92.2 & 97.2 &  93.0 \\
    \quad Chat Hard & 71.4 & 64.4 & 72.8 & \textbf{80.1} \\
    \quad Safety & 83.1 & 75.5 & 82.9 & \textbf{89.6} \\
    \quad Reasoning & 85.1 & 81.4 & 89.7 & \textbf{92.2} \\
    \bottomrule
    \end{tabular}}
    \vspace{-2mm}
\end{wraptable}
\paragraph{Evaluation, Results, and Analysis.} 
%Following the evaluation protocol of LE~\citep{zhang-etal-2025-lists}, we compare our method against three baselines: a standard Bradley–Terry (BT) model, BT† (trained on a filtered dataset with format-biased samples removed), and LE. 
As reported in Table~\ref{tab:res_format}, the standard BT model exhibits pronounced format bias, achieving win-rates of 89.0\% and 92.5\% for responses in Bold and List formats, respectively, providing strong evidence that vanilla BT conflates superficial formatting cues with response quality. The BT$\dag$ variant, while partially mitigating this bias through data filtering, suffers a substantial drop in downstream performance on RewardBench, indicating that naive removal of format-biased samples compromises the model’s ability to learn robust reward signals. In contrast, both FD and our method successfully neutralize format bias, driving win-rates close to the ideal 50\% threshold. Crucially, our approach outperforms FD on the more challenging subsets of RewardBench, demonstrating superior generalization in high-stakes domains, which underscores that our method achieves a more favorable trade-off between format debiasing and preference learning. % eliminating spurious format preferences without sacrificing and enhancing core capabilities essential for reliable reward modeling.

\vspace{-2mm}
\subsection{{Concurrent Multi-Biases}}
\vspace{-1mm}
% Real-world datasets frequently exhibit multiple concurrent biases. In this section, we investigate whether DIR can effectively mitigate such co-occurring biases—specifically, length bias and sycophancy bias—simultaneously. Our results demonstrate that the multi-bias variant of DIR substantially reduces both biases relative to the standard Bradley–Terry (BT) baseline. Notably,  , it even yields improved generalization performance, suggesting that incorporating multiple debiasing objectives into the training objective does not induce significant destructive interference during optimization. Full experimental details are provided in Appendix~\ref{app:concurrent_multi_bias}.
Real-world datasets frequently exhibit multiple concurrent biases. Hence, we investigate whether DIR can effectively mitigate such co-occurring biases, specifically, length bias and sycophancy bias, simultaneously. %Our results demonstrate that the multi-bias variant of DIR substantially reduces both biases relative to the standard Bradley–Terry (BT) baseline. Notably, when debiasing both length and sycophancy jointly, the resulting reward model achieves improved generalization performance on RM-Bench, indicating that the incorporation of multiple debiasing objectives into the training objective does not induce significant destructive interference during optimization. %Full experimental details are provided in Appendix~\ref{app:concurrent_multi_bias}.
Concretely, we extend DIR to length and sycophancy biases by introducing two independent mutual information terms, each with its own debiasing head $\psi_\text{length}$ and $\psi_\text{syco}$, respectively: %For each preference pair, we construct $\boldsymbol{b}_\text{rel}^\text{len}$ (which response is longer) and $\boldsymbol{b}_\text{rel}^\text{syco}$ (which response is more sycophantic), and optimize:
\begin{equation} \label{eq:final_loss_multi_bias}
\mathcal{L}_{\text{Total}}(\phi) = {\mathcal{L}_{\text{Preference}}}(\phi) + \lambda_{\text{length}} \cdot\mathcal{L}_{\text{Debiasing}}(\phi,\psi_\text{length}) + \lambda_{\text{syco}}\cdot \mathcal{L}_{\text{Debiasing}}(\phi,\psi_\text{syco}),
\end{equation}
where $\lambda_{\text{len}}$ and $\lambda_{\text{syco}}$ are set to $1.0$. We follow the setting in Table~\ref{tab:res_syco}, training Llama-3.1-8B-Instruct on the HelpSteer3 dataset under two sycophancy contamination configurations ($\gamma = 40\%$ and $80\%$, $ \alpha = 70\%$), where a larger $\gamma$ indicates a more challenging setting. All models (BT, length-only DIR, and length+syco DIR) are trained for 1 epoch on the same data, and we report results using the final checkpoint for fairness and convenience.

As shown in Table~\ref{tab:multi_bias_rm_bench}, on RM-Bench, the joint model (Ours-Len-Syco) achieves the best overall performance and the largest gains on the hardest subset (e.g., at $\gamma=40\%$, Total: $67.25 \rightarrow 69.96$, Hard: $39.88 \rightarrow 46.85$ vs. BT), while still clearly reducing the Pearson correlation with length relative to BT, confirming that length bias is mitigated even when sycophancy is also debiased. We also observe that the length-only model (Ours-Len) attains the lowest length–reward Pearson coefficient, but somewhat surprisingly, the joint length+sycophancy debiasing (Ours-Len-Syco) yields the best overall RM-Bench performance, suggesting that debiasing multiple biases together could also help lead to a more balanced reward model. As shown in Table~\ref{tab:multi_bias_syco}, on sycophancy stress tests, debiasing only sycophancy (Ours-Syco) gives the strongest syco robustness, as expected, but the joint model (Ours-Len-Syco) still substantially outperforms BT on all sycophancy metrics (All./Nat./Adv.) across both $\gamma$ settings, while additionally reducing length bias. In summary, a second debiasing term leads to a controlled trade-off, not conflicting gradients: both biases are improved over BT, and overall RM quality remains strong.

\begin{table}[t]
\centering
\caption{Performance comparison of concurrent multi-bias experiments on RM-Bench. Best results are in \textbf{bold}.}
\vspace{-2mm}
\resizebox{\textwidth}{!}{
\begin{tabular}{c|cccc|ccc|cc}
\toprule
$\gamma=40\%, \alpha=70\%$ & Chat & Math & Code & Safety & Hard & Normal & Easy & Total & Pearson Coefficient \\ \midrule
BT & 66.58 & 64.17 & 53.70 & 84.53 & 39.88 & 72.82 & \textbf{89.04} & 67.25 & 0.4807 \\
Ours-Len & 66.93  & 64.59 & 53.12 & \textbf{89.14} & 44.25 & 72.43 & 88.65 & 68.44 & \textbf{0.4235} \\
Ours-Len-Syco & \textbf{70.80} & \textbf{65.07} & \textbf{55.26} & 88.71 & \textbf{46.85} & \textbf{75.08} & 87.96 & \textbf{69.96} & 0.4446 \\ \midrule
$\gamma=80\%, \alpha=70\%$ & Chat & Math & Code & Safety & Hard & Normal & Easy & Total & Pearson Coefficient \\ \midrule
BT & 65.37 & 63.71 & 53.12 & 80.85 & 37.58 & 70.96 & \textbf{88.75} & 65.76 & 0.4666 \\
Ours-Len & \textbf{68.91} & 64.25 & 53.75 & 87.23 & 44.78 & \textbf{73.09} & 87.72 & 68.53 & \textbf{0.4081} \\
Ours-Len-Syco & 68.39 & \textbf{64.92} & \textbf{54.04} & \textbf{88.06} & \textbf{45.15} & 72.92 & 88.50 & \textbf{68.85} & 0.4235 \\ \bottomrule
\end{tabular}
}\label{tab:multi_bias_rm_bench}
\end{table}

\begin{table}[t]
\centering
\caption{Reward model accuracy (\%) on the concurrent multi-bias under different contamination settings.}
\vspace{-2mm}
\resizebox{\textwidth}{!}{
\begin{tabular}{cc|ccc|ccc|ccc}
\toprule
 &  & \multicolumn{3}{c}{All.} & \multicolumn{3}{c}{Nat.} & \multicolumn{3}{c}{Adv.} \\
$\gamma$ & $\alpha$ & BT & Ours-Syco & Ours-Len-Syco & BT & Ours-Syco & Ours-Len-Syco & BT & Ours-Syco & Ours-Len-Syco \\ \midrule
40\% & 70\% & 83.6 & 88.0 & 85.6 & 84.4 & 87.4 & 86.4 & 82.6 & 88.6 & 84.4 \\
80\% & 70\% & 81.2 & 86.2 & 85.9 & 86.4 & 87.2 & 86.6 & 79.7 & 86.2 & 85.1 \\ \bottomrule
\end{tabular}
}\label{tab:multi_bias_syco}
\vspace{-2mm}
\end{table}

\vspace{-2mm}
\section{Conclusion}
\vspace{-2mm}
%We introduce \textbf{D}ebiasing via \textbf{I}nformation optimization of \textbf{R}Ms (DIR), a novel reward model debiasing framework based on information-theoretic guidance. %designed to mitigate reward hacking caused by inductive biases in RLHF by applying information-theoretic principles to reward modeling. 
%
%Unlike prior works that target on single biases (e.g., length or format) or only address simple linear correlations (e.g., Pearson Coefficient), DIR directly confronts the root cause of reward hacking, inductive bias in preference data, by implementing a dual-objective to disentangle these signals explicitly. 

%DIR guides the reward model to learn representations that are predictive of true human preference while remaining invariant to the influence of known biases. Experiments across three distinct scenarios (i.e., length, sycophancy, and format bias) demonstrate DIR’s effectiveness not only in neutralizing the target biases but also in enhancing downstream RLHF performance and generalization, validating our approach as a general and practical tool for building more robustly aligned models.

We introduce \textbf{D}ebiasing via \textbf{I}nformation optimization of \textbf{R}Ms (DIR), a novel information-theoretic method, to address the pervasive issue of inductive biases in reward modeling for RLHF. By maximizing the mutual information between RM scores and genuine human preference signals while minimizing the mutual information between RM predictions and biased attributes, DIR effectively disentangles genuine human preference signals from spurious correlations, \textit{e.g.}, response length, sycophancy, and format. Equipped with variational bounds and MI estimation strategies, our method handles non-linear and complex bias structures with theoretical rigor and practical efficacy. Extensive experiments across diverse LLM and reward model benchmarks demonstrate that DIR not only mitigates a broad spectrum of biases but also enhances the generalization and alignment quality of reward models, leading to more robust downstream performance. We believe DIR offers a principled, scalable, and widely applicable solution for building more reliable and balanced alignment systems, paving the way toward more human-value-consistent artificial intelligence.
\bibliography{iclr2026_conference}
\bibliographystyle{iclr2026_conference}

\appendix
\clearpage

\section{Bound Proof}

\subsection{Proof of the Barber-Agakov (BA) Bound}
\label{app:proof_ba}
We aim to prove that for any variational distribution $q_\theta(\mathbf{y}|\mathbf{x})$, the mutual information $I(\mathbf{x}; \mathbf{y})$ is lower-bounded by:
\begin{equation}
    I(\mathbf{x}; \mathbf{y}) \geq \mathbb{E}_{p(\mathbf{x}, \mathbf{y})} [\log q_\theta(\mathbf{y}|\mathbf{x})] + H[p] =: I_{\text{BA}}(\mathbf{x}; \mathbf{y}),
\end{equation}
where $H[p]$ is the entropy of the ground-truth distribution $p(\vx,\vy)$. By the definition of mutual information, we have:
\begin{equation}
    I(\mathbf{x}; \mathbf{y}) = H(\mathbf{y}) - H(\mathbf{y}|\mathbf{x}) = H(\mathbf{y}) + \mathbb{E}_{p(\mathbf{x}, \mathbf{y})} [\log p(\mathbf{y}|\mathbf{x})].
\end{equation}
Consider the expected Kullback-Leibler (KL) divergence between the true conditional distribution $p(\mathbf{y}|\mathbf{x})$ and the variational approximation $q_\theta(\mathbf{y}|\mathbf{x})$, we have:
\begin{equation}\label{eq:ba_inequality}
    \mathbb{E}_{p(\mathbf{x})} \left[ D_{\text{KL}} \big( p(\mathbf{y}|\mathbf{x}) \,\|\, q_\theta(\mathbf{y}|\mathbf{x}) \big) \right] = \mathbb{E}_{p(\mathbf{x}, \mathbf{y})} \left[ \log \frac{p(\mathbf{y}|\mathbf{x})}{q_\theta(\mathbf{y}|\mathbf{x})} \right] \geq 0.
\end{equation}
By the linearity of expectation, the inequality~\ref{eq:ba_inequality} can be rearranged as:
\begin{equation}\label{eq:ba_rearranged}
    \mathbb{E}_{p(\mathbf{x}, \mathbf{y})} [\log p(\mathbf{y}|\mathbf{x})] \geq \mathbb{E}_{p(\mathbf{x}, \mathbf{y})} [\log q_\theta(\mathbf{y}|\mathbf{x})].
\end{equation}
Then we can substitute \eqref{eq:ba_rearranged} into the definition of $I(\mathbf{x}; \mathbf{y})$, yielding:
\begin{equation}
    I(\mathbf{x}; \mathbf{y}) \geq H[p] + \mathbb{E}_{p(\mathbf{x}, \mathbf{y})} [\log q_\theta(\mathbf{y}|\mathbf{x})],
\end{equation}
which completes the proof. The bound is tight if and only if $q_\theta(\mathbf{y}|\mathbf{x}) = p(\mathbf{y}|\mathbf{x})$ almost everywhere with respect to $p(\mathbf{x}, \mathbf{y})$.

\subsection{Proof of the CLUB Upper Bound}\label{app:proof_club}
We aim to prove that for any variational distribution $q_\theta(\vy|\vx)$, the mutual information $\MI{\vx}{\vy}$ is upper-bounded by ${I}_{\text{CLUB}}(\vx;\vy)$. We begin with the definition of mutual information:
\begin{align}
    \MI{\vx}{\vy} &= \bbE_{p(\vx,\vy)}\left[\log p(\vy|\vx)\right] - \bbE_{p(\vy)}\left[\log p(\vy)\right] \label{eq:club_proof_start}
\end{align}
Let's focus on the second term, which is the negative marginal entropy $+H(\vy)$. We can express the marginal distribution $p(\vy)$ by marginalizing out $\vx$:
\begin{equation}\label{eq:def_marginal_py}
    p(\vy) = \bbE_{p(\vx')}[p(\vy|\vx')]
\end{equation}
where $\vx'$ is a random variable drawn from the same distribution as $\vx$, but is independent of the $\vx$ in the first term of \eqref{eq:club_proof_start}. Substituting \eqref{eq:def_marginal_py} into the entropy term:
\begin{equation}
    - \bbE_{p(\vy)}\left[\log p(\vy)\right] = - \bbE_{p(\vy)}\left[\log \bbE_{p(\vx')}[p(\vy|\vx')]\right].
\end{equation}
Since the logarithm is a concave function, we can apply Jensen's inequality, which states that $\bbE[\log(Z)] \leq \log(\bbE[Z])$ and implies $-\log(\bbE[Z]) \leq -\bbE[\log(Z)]$. Applying this, we get:
\begin{equation}
    - \bbE_{p(\vy)}\left[\log \bbE_{p(\vx')}[p(\vy|\vx')]\right] \leq - \bbE_{p(\vy)}\left[\bbE_{p(\vx')}[\log p(\vy|\vx')]\right] = - \bbE_{p(\vx')p(\vy)}[\log p(\vy|\vx')].\label{eq:club_inqquality}
\end{equation}
Now, substituting \eqref{eq:club_inqquality} back into our original MI expression \eqref{eq:club_proof_start}, we obtain an upper bound on the mutual information:
\begin{equation}
    \MI{\vx}{\vy} \leq \bbE_{p(\vx,\vy)}[\log p(\vy|\vx)] - \bbE_{p(\vx)p(\vy)}[\log p(\vy|\vx)]. \label{eq:club_proof_p_bound}
\end{equation}
Note that the second expectation is over the product of marginals $p(\vx)p(\vy)$. The inequality \eqref{eq:club_proof_p_bound} holds for the true conditional distribution $p(\vy|\vx)$. The CLUB bound replaces $p(\vy|\vx)$ with the variational approximation $q_\theta(\vy|\vx)$. The key insight from \citet{cheng2020club} is that the difference between the true bound and the variational bound is an expectation of KL-divergences, and the proposed variational form serves as a practical, sample-based upper bound for minimization. Therefore, we use the variational form as our tractable objective:
\begin{equation}
    \MI{\vx}{\vy} \leq \bbE_{p(\vx,\vy)}[\log q_\theta(\vy|\vx)] - \bbE_{p(\vx)p(\vy)}[\log q_\theta(\vy|\vx)] =: {I}_{\text{CLUB}}(\vx;\vy),
\end{equation}
which completes the justification for using ${I}_{\text{CLUB}}$ as an upper bound for mutual information minimization.

% {The dimensions are chosen based on the following principles:
% input\_dim: This is dynamically set to match the dimension of the final hidden state representation from the backbone LLM. For instance, in our experiments with Llama-3.1-8B-Instruct, the input\_dim is 4096.}

% {label\_num: The output dimension is set to 2. This is a direct and necessary consequence of our theoretical formulation, as the relative bias attribute $\boldsymbol{b}_\text{rel}$ is defined as a binary variable ({0, 1}) indicating which of the two responses in a pair exhibits a stronger bias.}

% {hidden\_size: For the intermediate hidden layer size, we conducted an ablation study over the values [512, 1024, 2048]. We observed that hidden\_size=1024 offered the best trade-off between debiasing performance and minimal computational overhead.}

% {We intentionally designed $q_\psi$ to be a simple and efficient network. This ensures that the observed performance gains are attributable to our method itself, rather than to the introduction of a large number of additional parameters.}

\section{Experiment}
\subsection{Length Bias}\label{app:len_bias}

\paragraph{Training Settings.} %For our reward model training, we adopt a full parameter tuning strategy by using HuggingFace Trainer with DeepSpeed Zero1 on 8 GPU cards. Global batch size is set to 128, initialization learning rate $\alpha_1$ is $2e-6$ with Cosine scheduler. We set the learning rate $\alpha_2$ of the debiasing network as $1e-3$. 
For our PPO experiment, we fine-tune two distinct models using 20,000 samples from the alpaca-gpt4-data-en dataset~\citep{peng2023gpt4llm}. The first model, Llama3.1-8B-Instruct~\citep{grattafiori2024llama3herdmodels}, has undergone post-training that includes both DPO and RLHF. The second, OpenRLHF-Llama3-8B-SFT~\citep{dong2024rlhfworkflowrewardmodeling}, is an instruction-following version built upon Llama3-8B-Base, without the RLHF post-training stage. We conduct the PPO training using the ms-swift~\citep{zhao2025swiftascalablelightweightinfrastructure} framework with its default PPO training configuration. 

\paragraph{Baselines.} We mainly consider the following baselines due to the reproducibility: (1) Vanilla BT Baseline and popular open-source RM Skywork-Reward-Llama-3.1-8B-v0.2~\citep{liu2024skyworkrewardbagtricksreward}; (2) Length Debiased RMs, including PoE~\citep{shen2023loose} and ALBM~\citep{bu2025beyond}; (3) Length Penalty that directly resharps the reward during PPO by $\Tilde{r}(\vx,\vy)=r(\vx,\vy)-0.001*len(\vy)$~\citep{dong2024rlhfworkflowrewardmodeling}; (4) InfoRM~\citep{miao2024inform} that is also designed from the information theory perspective.

%\paragraph{Evaluations.} All benchmark evaluations are subsequently performed using the ms-evalscope framework. Our evaluation protocol utlize few-shot settings for GSM8K (4-shot)~\citep{cobbe2021trainingverifierssolvemath}, Race (3-shot)~\citep{lai2017racelargescalereadingcomprehension}, and TriviaQA (5-shot)~\citep{joshi2017triviaqalargescaledistantly}, while all other benchmarks (i.e., Hellaswag~\citep{zellers2019hellaswagmachinereallyfinish}, IFeval~\citep{zhou2023instructionfollowingevaluationlargelanguage}, MMLU~\citep{hendrycks2021measuringmassivemultitasklanguage}, ProcessBench~\citep{zheng2025processbenchidentifyingprocesserrors}, BBH~\citep{suzgun2022challengingbigbenchtaskschainofthought}, and HumanEval~\citep{chen2021evaluatinglargelanguagemodels}) are assessed in a zero-shot setting. We report accuracy as the primary metric for all tasks, with the exception of HumanEval, for which we report the Pass@1 score.

\paragraph{Performance on RM-Bench.} 
We further evaluate our debiased reward models on {RM-Bench}~\citep{liu2024rmbenchbenchmarkingrewardmodels}, a comprehensive benchmark that assesses model capabilities across four distinct domains, including \texttt{Chat}, \texttt{Math}, \texttt{Code}, and \texttt{Safety} with three difficulty levels: \texttt{Hard}, \texttt{Normal}, and \texttt{Easy}. As shown in Table~\ref{tab:perf_on_rmbench}, our DIR framework consistently outperforms several strong baselines in terms of overall performance. Our primary variant, denoted as {Ours-1.0}, corresponds to the optimal trade-off point identified in our ablation study ($\lambda = 1.0$), which achieves the second-highest aggregate score of {69.35}, reflecting a well-calibrated balance between debiasing and generalization and indicating that DIR enhances the reward model’s discriminative capacity on core reasoning tasks without substantially degrading its general-purpose alignment.

When we increase the debiasing strength to $\lambda = 10.0$, the resulting model {Ours-10.0} achieves the highest overall score of {70.18}. The most pronounced improvement occurs on the {Hard} subset, where performance surges to {64.41}, surpassing the next-best method by over {16 points}. Such a substantial performance improvement suggests that by explicitly suppressing reliance on superficial bias through the DIR mechanism, the reward model is better able to attend to nuanced, content-based indicators of response quality, particularly those that are critical for evaluating complex or challenging prompts. Moreover, {Ours-10.0} achieves the top scores in both the \texttt{Chat} and \texttt{Code} domains. However, this stronger debiasing comes at a cost: performance on the \texttt{Easy} subset declines relative to weaker debiasing settings. On such instances, where simple heuristics often suffice for accurate judgment, the aggressive removal of bias signals appears overly restrictive and counterproductive.  In summary, these results demonstrate that DIR not only enhances the overall capability of the reward model but also offers a tunable mechanism to prioritize robustness on challenging tasks over simpler ones, howcasing the flexibility and effectiveness of our approach.

\begin{table*}[h!]
\centering
\caption{Performance comparison on RM-Bench. Best results are in \textbf{bold}. Second-performance is \underline{underlined}.}
\label{tab:perf_on_rmbench}
\resizebox{0.76\textwidth}{!}{%
\begin{tabular}{c|cccc|ccc|c}
\toprule
Method & Chat & Math & Code & Safety & Hard & Normal & Easy & Total \\ \midrule
BT & 64.69 & 61.21 & 51.41 & 95.11 & 42.76 & 72.30 & \underline{89.24} & 68.10 \\
PoE & 67.70 & 61.23 & 51.51 & \textbf{95.51} & 44.94 & \underline{73.17} & 88.86 & 68.99 \\
ALBM & 64.57 & 58.48 & \underline{52.34} & \underline{95.21} & 47.88 & 71.50 & \textbf{90.32} & 67.40 \\ \midrule
Ours-1.0 & \underline{68.91} & \textbf{61.81} & 51.56 & 95.13 & \underline{47.88} & \textbf{73.59} & 88.93 & \underline{69.35} \\
Ours-10.0 & \textbf{71.23} & \underline{61.59} & \textbf{52.73} & 94.91 & \textbf{64.41} & 71.29 & 74.85 & \textbf{70.18} \\ \bottomrule
\end{tabular}
}
\end{table*}

\paragraph{Performance on MT-Bench and AlpaceEval.}\label{app:performance_mtben_alpaca}
{For MT-Bench~\citep{zheng2023judgingllmasajudgemtbenchchatbot}, we report the win rate of each RM-guided policy against its own base model, using the standard MT-Bench LLM-as-a-judge setup. As shown in Table~\ref{tab:res_mtbench}, our method (``Ours'') achieves the highest win rates on both backbones (56.25\% vs. 48.75–53.75\% for OpenRLHF-Llama-3-8B-SFT, and 56.88\% vs. 50.63–51.88\% for Meta-Llama3.1-8B-Instruct), indicating more improvements on open-ended, multi-turn dialogue quality.}

\begin{table}[t]
\centering
\caption{Win rate (\%) performance comparison on MT-Bench.}
\resizebox{0.7\textwidth}{!}{
\begin{tabular}{c|cc}
\toprule
Win Rate (\%)  & \multicolumn{2}{c}{Base Model} \\ 
(vs. Base) & OpenRLHF-Llama-3-8B-SFT & Meta-Llama3.1-8B-Instruct\\ \midrule
Ours & \textbf{56.25} &  \textbf{56.88}  \\
PoE & 48.75 &  51.25  \\
Skywork & 49.38  & 51.25  \\
ALBM & 53.75 &  50.63  \\
InfoRM & 46.88 & 51.88 \\ \bottomrule
\end{tabular}
}\label{tab:res_mtbench}
\end{table}

{For Length Controlled AlpacaEval, we follow the length-controlled protocol of~\citet{dubois2025lengthcontrolledalpacaevalsimpleway} and report both raw win rate and length-controlled win rate over the base model. On Meta-Llama3.1-8B-Instruct, Ours achieves the highest scores on both metrics. On OpenRLHF-Llama-3-8B-SFT, Skywork attains a slightly higher raw win rate, but Ours achieves the best length-controlled win rate, which is consistent with our goal: once the confounding effect of response length is controlled for, our debiased RMs yield policies that are preferred more often, demonstrating better alignment that is not driven by verbosity. We will include these MT-Bench and AlpacaEval results and their analysis in the revised version.}

\begin{table}[t]
\centering
\caption{Performance comparison on Length Controlled AlpacaEval against \texttt{gpt4-1106-preview}.}
\resizebox{0.6\textwidth}{!}{
\begin{tabular}{ccc}
\toprule
\multicolumn{3}{c}{Base Model: Meta Llama3.1-8B-Instruct} \\
Methods & Raw Win Rate (\%) & Length Control Win Rate (\%) \\ \midrule
Ours & \textbf{31.30} & \textbf{19.66} \\
PoE & 26.58 & 11.41 \\
Skywork & 29.38 & 13.21 \\
ALBM & 26.83 & 10.61 \\
InfoRM & 25.22 & 11.02 \\ \midrule
\multicolumn{3}{c}{Base Model: OpenRLHF Llama-3-8B-SFT} \\
Methods & Raw Win Rate (\%) & Length Control Win Rate (\%) \\ \midrule
Ours & 9.50 & \textbf{5.46} \\
PoE & 7.14 & 3.28 \\
Skywork & \textbf{10.19} & 3.93 \\
ALBM & 8.88 & 5.08 \\
InfoRM & 5.84 & 3.65 \\ \bottomrule
\end{tabular}
}\label{tab:res_alpacaeval2}
\end{table}

\begin{figure}[t]
    
    \centering
    \includegraphics[width=1\linewidth]{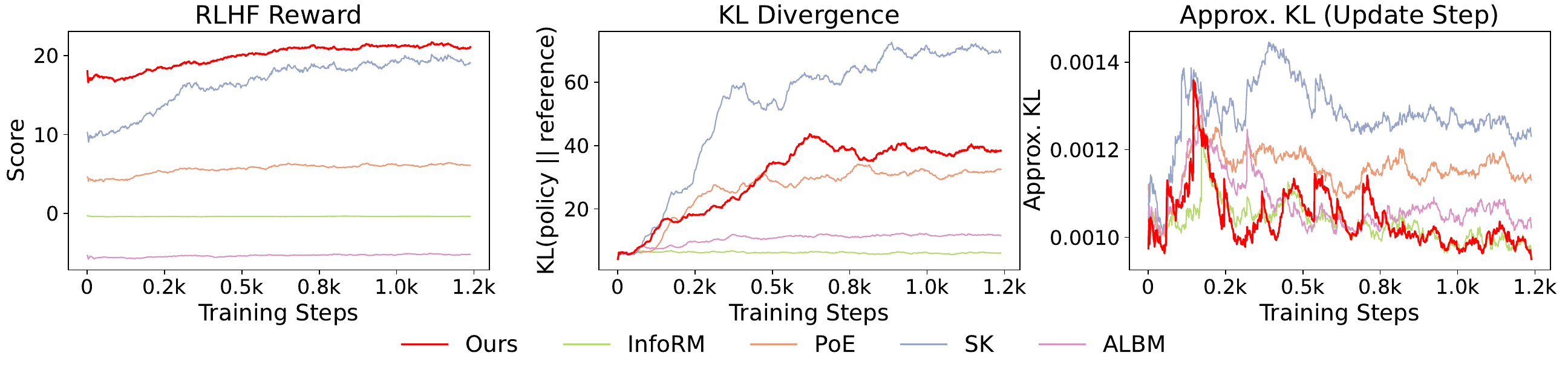}
    \caption{PPO training dynamics across key metrics. Our RM obtains a higher policy score and demonstrates better training stability.}
    \label{fig:ppo_vis}
\end{figure}

\paragraph{PPO Training Monitoring.} Figure~\ref{fig:ppo_vis} presents three key metrics for monitoring the PPO training process. The left plot (RLHF Reward) evaluates the final quality score of the model's outputs, with higher values being better. The middle plot (KL Divergence) measures how much the learned policy has deviated from the initial reference model, indicating the extent of exploration. The right plot (Approx. KL) shows the magnitude of each policy update, serving as a critical indicator of training stability. Our policy model demonstrates a better balance across these metrics by achieving a top reward score that significantly outperforms all baselines. Concurrently, our KL divergence is maintained at a moderate level, suggesting effective exploration without catastrophic deviation from the base model's capabilities. Most importantly, our method exhibits the lowest and most stable Approx. KL, which proves that the training process is exceptionally smooth and reliable. In summary, our approach successfully boosts performance while ensuring unparalleled training stability.

\subsection{RM Training Cost Analysis.}\label{app:training_cost}
We analyze the computational overhead in terms of GPU memory consumption and training time, with a detailed comparison presented in Table \ref{tab:trainimg_cost}. We use 8 GPU cards with full parameter training and DeepSpeed Zero-1\citep{rajbhandari2020zeromemoryoptimizationstraining}. Our approach demonstrates highly comparable resource efficiency to existing methods. Specifically, the GPU memory usage of our method (57.22GB) is only marginally higher than the baseline (56.80GB) and on par with other techniques like ALBM (56.88GB). Regarding training time, while our method (67.09 minutes) requires a moderate increase compared to the simpler baseline (50.46 minutes), DIR remains competitive and aligns closely with other advanced methods such as ALBM (68.21 minutes), which shows that the significant performance improvements offered by our approach do not come at the expense of prohibitive computational costs, establishing our approach as a practical and efficient solution.
\begin{table}[t]
\centering
\caption{Training cost comparison.}
\resizebox{0.4\textwidth}{!}{
\begin{tabular}{c|cc}
\toprule
Method & GPU Memory & Training Time \\ \midrule
Baseline & 55.08GB & 50.46m \\
PoE & 56.80GB & 55.35m \\
ALBM & 57.22GB & 78.21m \\
InfoRM & 57.99GB & 75.21m \\
Ours & 56.88GB & 67.09m\\ \bottomrule
\end{tabular}
}\label{tab:trainimg_cost}
\end{table}

\subsection{Ablation Studies under Length Debias}\label{app:ablation_length}
\paragraph{Ablation Study on Representation for Debiasing.}
In this section, we investigate the influence of difference-based representation  $\Delta\vh = \vh^w - \vh^l$ as input to the variational debiasing network $q_\psi$, compared with the representation of the concatenating form $[\vh^w; \vh^l]$. We compare both variants on \texttt{RewardBench-v1} and \texttt{RM-Bench}. Results in Table~\ref{tab:ablation_repr} show that the difference-based variant slightly outperforms the concatenation-based one across most domains and difficulty levels. Empirically, the difference-based approach yields clear gains. On \texttt{RewardBench-v1}, accuracy on \texttt{Chat Hard} improves from 78.9\% to 83.6\%, and on \texttt{Reasoning} from 88.8\% to 90.0\%. On \texttt{RM-Bench}, the \texttt{Chat} score increases from 63.9\% to 66.8\%. Performance on other subsets remains stable, indicating no trade-off in generalization. From an efficiency standpoint, the difference operator preserves the embedding dimensionality, whereas concatenation doubles it. The reduced input size lowers both parameter count and GPU memory usage during training. Given its empirical advantage, theoretical grounding, and computational efficiency, we adopt representation difference as the default input formulation in DIR.

\begin{table*}[h!]
\centering
\caption{Ablation study on the representation format for the debiasing module. We report accuracy (\%) on RewardBench-v1 and RM-Bench. The difference-based approach consistently outperforms concatenation, especially on challenging conversational and reasoning tasks. Best results are in \textbf{bold}.}
\label{tab:ablation_repr}
\resizebox{0.85\textwidth}{!}{%
\begin{tabular}{lcccccccc}
\toprule
& \multicolumn{4}{c}{{RewardBench-v1 (Acc \%)}} & \multicolumn{4}{c}{{RM-Bench (Acc \%)}} \\
\cmidrule(lr){2-5} \cmidrule(lr){6-9}
{Method} & Chat & Chat Hard & Safety & Reasoning & Chat & Math & Code & Safety \\
\midrule
Concat ($[\vh^w; \vh^l]$) & 93.3 & 78.9 & \textbf{90.9} & 88.8 & 65.9 & 60.8 & \textbf{52.6} & 95.0 \\
{Difference} ($\Delta\vh$) & \textbf{94.1} & \textbf{83.6} & 89.7 & \textbf{90.0} & \textbf{67.8} & \textbf{61.1} & 52.4 & \textbf{95.2} \\
\bottomrule
\end{tabular}%
}
\end{table*}

\begin{figure}[h!]
    \centering
    \includegraphics[width=0.8\columnwidth]{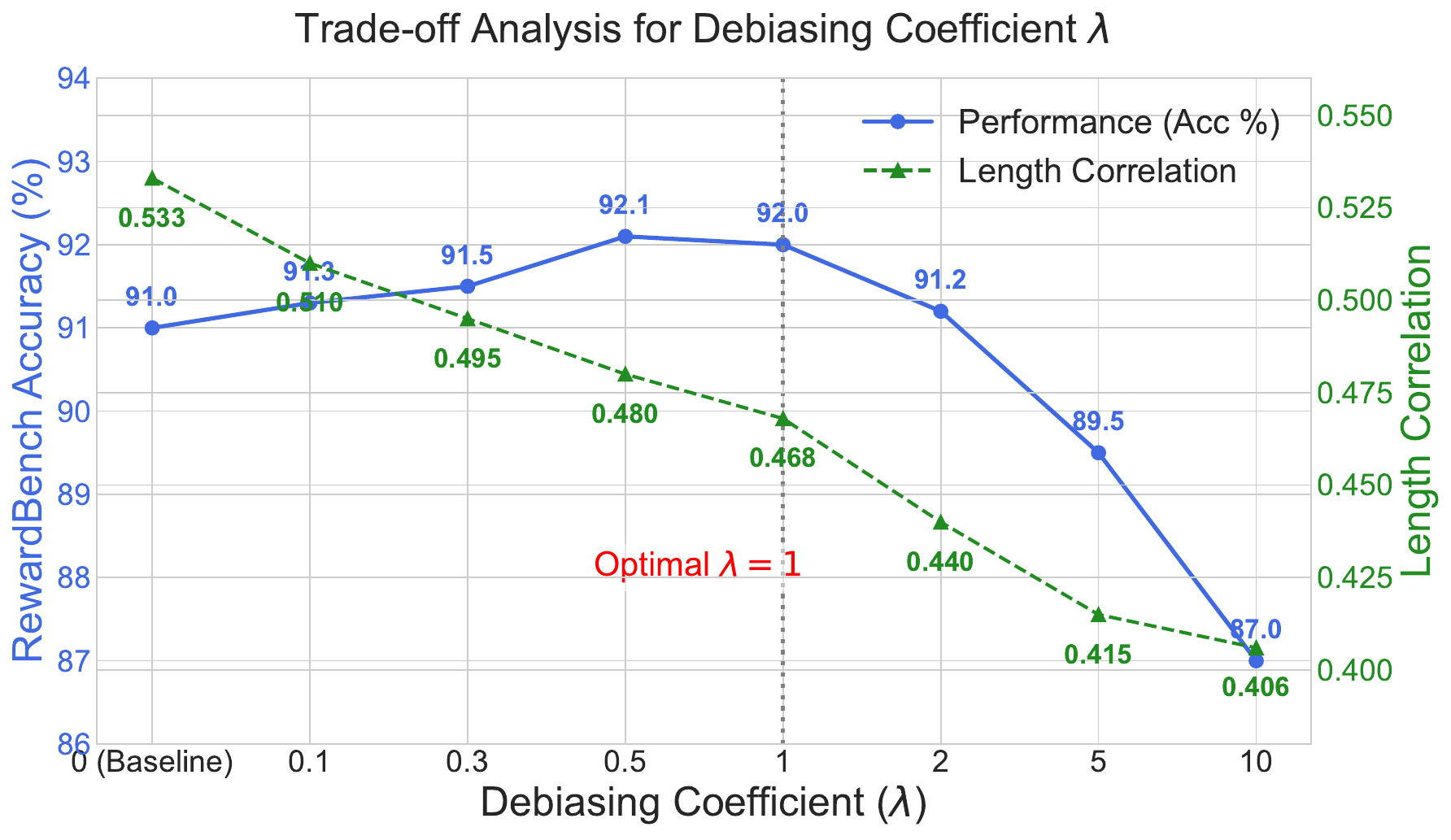}
    \caption{Ablation study on the debiasing coefficient $\lambda$. The plot shows the trade-off between preference learning performance (RewardBench Accuracy, blue) and the bias metric (e.g., Pearson $r$, green). $\lambda=1$ achieves the best balance.}
    \label{fig:ablation_lambda}
\end{figure}

\paragraph{Ablation Study on Debiasing Coefficient $\lambda$.}
The hyperparameter $\lambda$ in Equation~\ref{eq:final_loss} governs the trade-off between the standard preference learning objective ($\mathcal{L}_{\text{Preference}}$) and our information-theoretic debiasing objective ($\mathcal{L}_{\text{Debiasing}}$). To analyze its sensitivity, we tested a range of values: $\lambda \in \{0.1, 0.3, 0.5, 1, 2, 5, 10\}$. The results, visualized in Figure~\ref{fig:ablation_lambda}, reveal a clear trade-off.

As shown in the figure, when $\lambda$ is too small (e.g., $0.1$), the debiasing signal is insufficient. The model behaves similarly to a standard BT model, exhibiting a high bias metric (e.g., high Pearson correlation with a bias attribute) while achieving good performance on RewardBench. Conversely, when $\lambda$ is too large (e.g., $10$), the debiasing objective dominates the training. The resulting ``over-correction'' successfully minimizes the bias metric but severely compromises the reward model's ability to learn true preference signals, leading to a significant drop in RewardBench accuracy. Moreover, we observe that $\lambda = 1$ strikes an optimal balance. At $\lambda = 1$, the bias metric is substantially reduced, while the preference learning performance on RewardBench is maximized. The behavior demonstrates that the proposed method can effectively neutralize spurious correlations without damaging—and in fact while enhancing—the reward model's core capabilities. Therefore, all main experiments in this paper use $\lambda = 1$.

\subsection{Experiment on DPO}\label{app:dpo_exp}
Specifically, we adopt the ms-swift framework with its default DPO training configuration on Human-Like-DPO-Dataset~\citep{calik2025enhancinghumanlikeresponseslarge}, based on both OpenRLHF-Llama-3-8b-SFT and Meta-Llama3.1-8B-Instruct models, where DPO $\beta=0.1$, debias factor $\lambda=1$. We train 1 epoch and evaluate the performance on the final checkpoint. Human-Like-DPO-Dataset is created to fine-tune LLMs toward generating more human-like responses, which includes 10,884 samples across 256 topics, covering technology, daily Life, science, history and arts. We evaluate the performance on ArenaHard-v0.1 and several popular benchmarks. For baselines, we also compare with the length-controlled DPO method~\citep{park-etal-2024-disentangling}, which disentangles the length from the quality to explicitly avoid the policy model from preferring the longer response DPO training. 

%\subsection{Sycophancy Bias}\label{app:syco_bias}

%\subsection{Format Bias}\label{app:format_bias} 

%\subsection{Concurrent Multi-Bias}\label{app:concurrent_multi_bias}

\section{Prompt-based Justification Prompt}
We provide a Qwen3-235B-A22B-based pair-wise justification prompt shown below, which is adopted from ArenaHard's official implementation.

\begin{tcolorbox}
\begin{flushleft}
\textit{
Please act as an impartial judge and evaluate the quality of the responses provided by two AI assistants to the user prompt displayed below. You will be given assistant A's answer and assistant B's answer. Your job is to evaluate which assistant's answer is better. Begin your evaluation by generating your own answer to the prompt. You must provide your answers before judging any answers. When evaluating the assistants' answers, compare both assistants' answers with your answer. You must identify and correct any mistakes or inaccurate information. Then consider if the assistant's answers are helpful, relevant, and concise. Helpful means the answer correctly responds to the prompt or follows the instructions. Note that when a user prompt has any ambiguity or more than one interpretation, it is more helpful and appropriate to ask for clarifications or more information from the user than providing an answer based on assumptions. Relevant means all parts of the response closely connect or are appropriate to what is being asked. Concise means the response is clear and not verbose or excessive. Then consider the creativity and novelty of the assistant's answers when needed. Finally, identify any missing important information in the assistants' answers that would be beneficial to include when responding to the user prompt. After providing your explanation, you must output only one of the following choices as your final verdict with a label: \\
\hspace*{\fill}\\
1. Assistant A is significantly better: [[A $>>$ B]] \\
2. Assistant A is slightly better: [[A$>$B]] \\
3. Tie, relatively the same: [[A$=$B]] \\ 
4. Assistant B is slightly better: [[B$>$A]] \\ 
5. Assistant B is significantly better: [[B$>>$A]] \\
Example output: ``My final verdict is tie: [[A$=$B]]'' \\
\hspace*{\fill} \\ 
\text{[User Prompt]} \\ 
\hspace*{\fill} \\
\text{[Assistant A's Answer]} \\  
\text{\{answer A\}} \\
\hspace*{\fill} \\
\text{[Assistant B's Answer]} \\ 
\text{\{answer B\}} \\ 
}
\end{flushleft}
\end{tcolorbox}

\section{Case Study}\label{app:case_study}
\begin{figure}[t]
    \centering
    \includegraphics[width=0.75\linewidth]{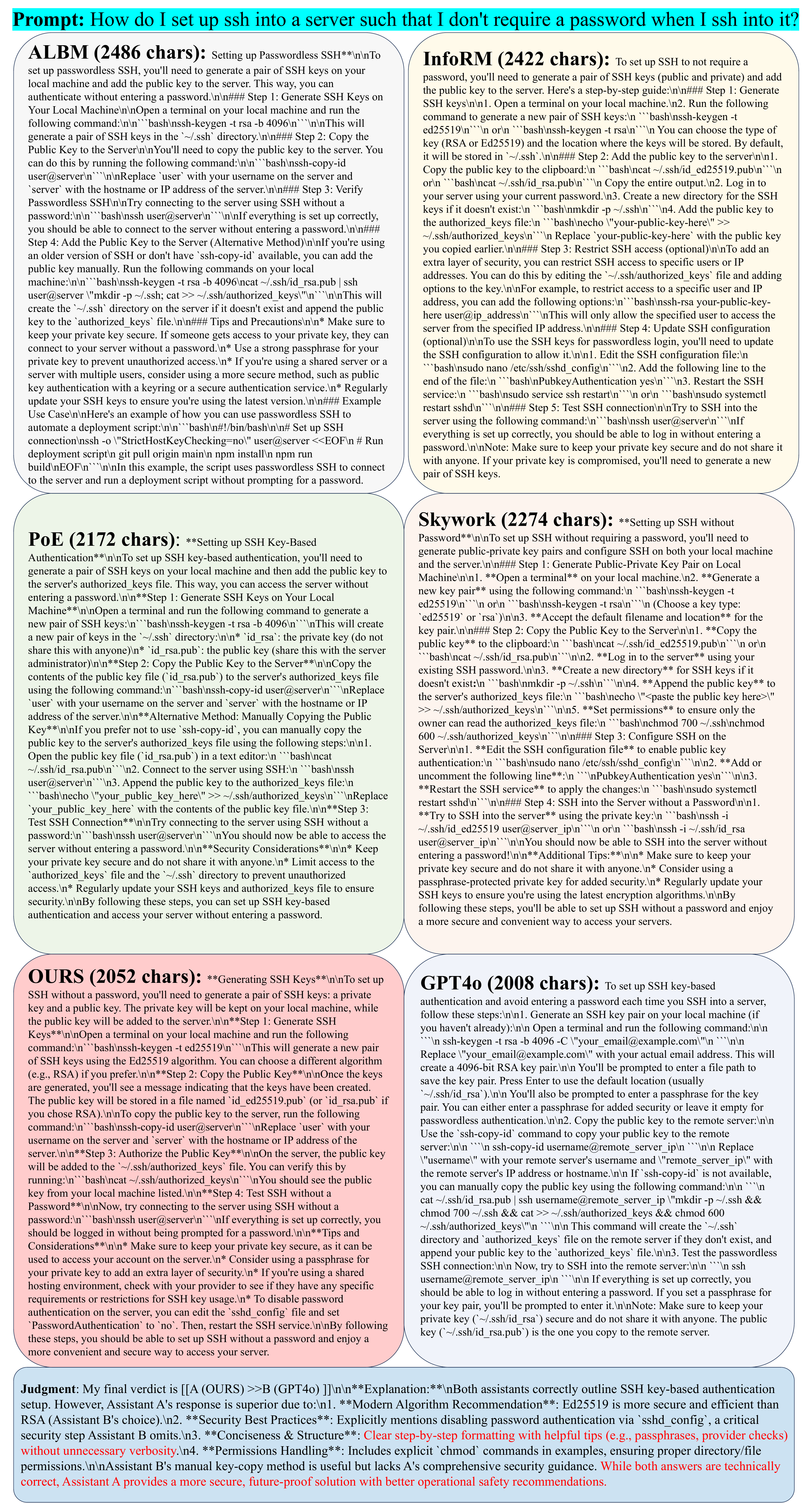}
    \caption{A case study on ArenaHard-v0.1 Benchmark where our model's response significantly outperforms GPT-4o on a technical question about setting up passwordless SSH.}
    \label{fig:case_study_1}
\end{figure}

Our model demonstrates its ability to generate not just a correct, but a more professionally helpful response compared to strong baselines like GPT-4o. As visualized in Figure~\ref{fig:case_study_1} (Data UID: \texttt{95bb5011680c4ba281ad7830ffa57f92}), when asked how to set up passwordless SSH, our model provided a guide that was praised by the LLM-based judge for its clear structure and organization. In this instance, GPT-4o's response is judged to be better to all other baselines except for our response, which provides a more secure, modern, and concise solution, highlighting its advanced capabilities.

Crucially, the judge noted that our model successfully avoids unnecessary verbosity, resulting in a response that is of higher quality and more concise. This is not merely a stylistic advantage; the substance of our model's response was also technically useful, which recommended the more modern and secure ``ed25519'' encryption algorithm over the RSA suggested by GPT-4o. Furthermore, our model included the critical security best practice of disabling password authentication entirely via the sshd\_config file—a vital step for server hardening that GPT-4o and other models omitted.

\end{document}